\documentclass[a4,12pt]{article}
\usepackage[numbers,sort,compress]{natbib}
\usepackage[utf8]{inputenc}
\usepackage[british]{babel}
\usepackage{libertine}
\usepackage[T1]{fontenc}
\usepackage{geometry}
\usepackage{hyperref}
\usepackage{caption}
\hypersetup{
    colorlinks=true,
    linkcolor=blue,
    filecolor=magenta,
    urlcolor=blue,
}
\usepackage{cuted}
\usepackage{lipsum}
\usepackage{graphicx}
\usepackage{enumerate}
\usepackage{amsmath}
	\numberwithin{equation}{section}
\usepackage{amssymb}

\usepackage{amsthm}
	\theoremstyle{plain}

	\theoremstyle{definition}

	\theoremstyle{remark}
		\newtheorem{rem}{Remark}

\usepackage{mathtools} % used to define := and =:
\usepackage{mathrsfs} % math script
\usepackage{eucal} % a better mathcal
\usepackage{braket} % for the command \Set: writes properly sets as \left\{ your elements \right\}
\usepackage{microtype}
\usepackage{mparhack}
\usepackage{booktabs}	% for better tables
\usepackage{siunitx}	% to measure units and tables with completely numeric columns
\usepackage{subfig}
\usepackage{listings}
\usepackage{algorithm}
\usepackage{algpseudocode}
\usepackage[labelfont=bf]{caption}
\usepackage[svgnames, table]{xcolor}
\usepackage{tikz}
	\usetikzlibrary{tikzmark,calc}
		
	\algnewcommand\algorithmicinput{\textbf{Input:}}
	\algnewcommand\INPUT{\item[\algorithmicinput]}
	\algnewcommand\algorithmicoutput{\textbf{Output:}}
	\algnewcommand\OUTPUT{\item[\algorithmicoutput]}
	\algdef{SE}[FUNCTION]{Function}{EndFunction}[2]{\algorithmicfunction\ \textproc{#1}\ifthenelse{\equal{#2}{}}{}{(#2)}}{\algorithmicend\ \algorithmicfunction}

\usepackage{tcolorbox}
\usepackage{titling}
\usepackage{titletoc}
\usepackage{longtable}
\usepackage{lscape}
\usepackage{capt-of}
\usepackage{marvosym}
\DeclareFixedFont{\ttb}{T1}{txtt}{bx}{n}{12} % for bold
\DeclareFixedFont{\ttm}{T1}{txtt}{m}{n}{12}  % for normal
\definecolor{deepblue}{rgb}{0,0,0.5}
\definecolor{deepred}{rgb}{0.6,0,0}
\definecolor{deepgreen}{rgb}{0,0.5,0}
\newcommand\pythonstyle{\lstset{
language=Python,
basicstyle=\ttm,
otherkeywords={self},             % Add keywords here
keywordstyle=\ttb\color{deepblue},
emph={MyClass,__init__},          % Custom highlighting
emphstyle=\ttb\color{deepred},    % Custom highlighting style
stringstyle=\color{deepgreen},
frame=tb,                         % Any extra options here
showstringspaces=false,
tabsize=2         %
}}
\lstnewenvironment{python}[1][]
{
\pythonstyle
\lstset{#1}
}
{}
\newcommand\pythoninline[1]{{\pythonstyle\lstinline!#1!}}

\renewcommand{\leq}{\leqslant}
\renewcommand{\geq}{\geqslant}
\newcommand{\ie}{i.\,e.~} % id est
\newcommand{\eg}{e.\,g.~} % exemplum gratia
\newcommand{\argmax}{\mbox{argmax}}

\newcommand{\softmax}{\mbox{softmax}}
\usepackage[intoc]{nomencl}

\makenomenclature

\makeatletter
\newcommand{\verbatimfont}[1]{\renewcommand{\verbatim@font}{\ttfamily#1}}
\makeatother
\usepackage{appendix}
\usepackage{cleveref}
\usepackage{wrapfig}
\crefname{thm}{Theorem}{Theorems}
\crefname{rem}{Remark}{Remarks}
\crefname{example}{Example}{Examples}
\crefname{paragraph}{Paragraph}{Paragraphs}
\crefname{appsec}{Supplementary Material}{Supplementary Material}
\crefname{appfig}{Supplementary Figure}{Supplementary Figures}
\crefname{apptab}{Supplementary Table}{Supplementary Tables}
\usepackage{authblk}
\title{idtracker.ai: Tracking all individuals in large collectives of unmarked animals}
\author[1,2]{Francisco Romero-Ferrero}
\author[1,2]{Mattia G. Bergomi}
\author[1]{Robert Hinz}
\author[1]{Francisco J. H. Heras}
\author[1,3]{Gonzalo G. de Polavieja}
\affil[1]{Champalimaud Research, Champalimaud Center for the Unknown - Lisbon, Portugal}
\affil[2]{F.R-F. and M.G.B. contributed equally to this work.}
\affil[3]{Correspondence for materials: gonzalo.polavieja@neuro.fchampalimaud.org}
\date{\today}
\begin{document}
\maketitle
\textbf{Our understanding of collective animal behavior is limited by our ability to track each of the individuals. We describe an algorithm and software, idtracker.ai, that extracts from video all trajectories with correct identities at a high accuracy for collectives of up to 100 individuals. It uses two deep networks, one detecting when animals touch or cross and another one for animal identification, trained adaptively to conditions and difficulty of the video.} \newline
\newline
Obtaining animal trajectories from a video faces the problem of how to track animals with correct identities after they touch, cross or they are occluded by environmental features. To bypass this problem, we proposed in idTracker the idea of tracking by identification of each individual using a set of reference images obtained from the video \cite{Perez-Escudero2014}. idTracker and further developments in animal identification algorithms \cite{Dolado2015,Rasch2016,Rodriguez2017,Wang2017,XU2017} can work for small groups of $2$-$15$ individuals. In larger groups, they only work for particular videos with few animal crossings \cite{Lecheval2017} or with few crossings of particular species-specific features \cite{Wang2017}.

Here we present idtracker.ai, a system to track all individuals in small or large collectives (up to $100$ individuals) at a high identification accuracy, often of $>99.9\%$. The method is species-agnostic and we have tested it in small and large collectives of zebrafish, \textit{Danio rerio} and flies, \textit{Drosophila melanogaster}. Code, quickstart guide and data used are provided (see \textbf{Methods}), and \textbf{Supplementary Text} describes algorithms and gives pseudocode. A graphical user interface walks users through tracking, exploration and validation (\textbf{Fig. 1a}).

Similar to idTracker \cite{Perez-Escudero2014}, but with different algorithms, idtracker.ai identifies animals using their visual features. In idtracker.ai, animal identification is done adapting deep learning \cite{LeCun2015,Abadi,Rusk2015} to work in videos of animal collectives thanks to specific training protocols. In brief, it consists of a series of processing steps summarized in \textbf{Fig. 1b}. After image preprocessing, the first deep network finds when animals are touching or crossing. Then the system uses the images between these detected to train a second deep network for animal identification. The system first assumes that a single portion of video when animals do not touch or cross has enough images to properly train the identification network (Protocol 1). However, animals touch or cross often and this portion is then typically very short, making the system estimate that identification quality is too low. If this happens, two extra Protocols (Protocols 2 and 3) are dedicated to safely accumulate enough images of each animal using several of these portions of video and build a larger training set. After training and assignation of identities, some postprocessing is performed to output the trajectories and an estimation of identification accuracy.

In the following we give more details of the processing steps. The preprocessing extracts blobs, areas of each video frame
corresponding to either a single animal, or to several animals that are touching or crossing, i.e. `crossings'. Then,
it orients the blobs using their axes of maximum elongation (\textbf{Fig. 1c}). This procedure leaves the animal pointing in one of two possible orientations.
We solve this ambiguity by training
with both upright and 180-degrees rotated images. This method is valid
for any elongated animal and is preferred to species-specific methods.

The deep
\textit{crossing detector} network finds whether each preprocessed image corresponds to a single animal or a crossing (\textbf{Fig. 1d}; details of network architecture in \textbf{Supplementary Table 1}). idtracker.ai
trains this network using images that confidently classifies as
single animals or crossings (see \textbf{Supplementary Text} for heuristics used). Once trained, it classifies all blobs as single animals or crossings. We depict these detected crossings as small black segments in \textbf{Fig. 1g}.

The deep \textit{identification}
network is then used to identify each individual between two crossings
(\textbf{Fig. 1e}; \textbf{Supplementary Table 1} for details of network architecture). We measured the
identification capacity of this network using $184$ single-animal videos, with $300$ pixels per animal on average. The advantage of single-animal videos is that we obtain a very large number of images per animal. Out of the $18,000$ images per animal we randomly selected $3,000$ for training. Testing it in $300$ new images gave a $>95 \%$
single-image accuracy up to $150$ animals (\textbf{Fig. 1f}; see \textbf{Supplementary
Fig. 2} for experimental set-up, \textbf{Supplementary Fig. 1} for results using alternative architectures detailed in \textbf{Supplementary Tables 2-3}). In contrast, idTracker degrades more quickly down to a value $\approx 83 \%$ for $30$ individuals and it is computationally too demanding for larger groups.

In videos of collective animal behavior, however, we lack direct access to $3,000$ images per animal to train the \textit{identification network}. Instead, we use a cascade of three protocols that obtains the training images differently depending on the difficulty of the video (\textbf{Fig. 1b}, cascade of training protocols; see \textbf{Supplementary Figures 3-4} for setups of video acquisition in zebrafish, and flies, respectively).

Protocol 1 starts by finding all intervals of the video where all the animals are detected as separated from each other. To each interval, for each animal we add images up to the next crossing from future frames, and images up to the immediate previous crossing from past frames.
We call \textit{global fragments} these extended intervals, which can contain different number of images per animal. Among all the global fragments, the system then chooses the one in which the animal traveling the shortest distance travels more (\textbf{Fig. 1g}, Step 1, colors indicate each of the $100$ individuals in the collective).
The system uses this global fragment to train the \textit{identification network}. Once trained, the network assigns identities in all the remaining global fragments.

Afterwards, the system evaluates the quality of the assigned global fragments. It eliminates: 1. global fragments with an estimated identification accuracy below some threshold, 2. those with identifications inconsistent with already assigned global fragments, and 3. those where the same identity has been assigned to several animals.
If the remaining high-quality global fragments (\textbf{Fig. 1g}, Step 2) cover $<99.95$\% of the images in global fragments, then Protocol 1 failed and Protocol 2 starts, as in our example.

Protocol 2 starts by training the network with the high-quality global fragments found in Protocol 1. This network is then used to assign the remaining global fragments again, selecting those passing our three-steps quality check. This procedure iterates until we have at least $50\%$ of images assigned.
From this point on, the system runs the accumulation as before, alternating it with the following extension. Single-animal fragments belonging to an unsuitable global fragment are accumulated if they are certain enough, are consistent with fragments already accumulated and do not introduce identity duplications.
Accumulation continues until no more acceptable global fragments remain or $99.95\%$ of the images from global fragments have survived the quality check. After this point, if $>90\%$ of the of the images in global fragments have been accumulated, then Protocol 2 ends successfully.
In our example, Protocol 2 stops the accumulation at the $9$th step (\textbf{Fig. 1g}, Step 9). Afterwards, the remaining images are assigned using the final network (see higher transparency segments in the close-up given in \textbf{Fig. 1h}).

The system then estimates identification accuracy using a conservative Bayesian framework (\textbf{Supplementary Fig. 5}), $99.95 \%$ in our example (\textbf{Fig. 1i}, top). Human validation of $3,000$ sequential video frames, by revising $680$ crossings, gave $99.997 \%$ (\textbf{Fig. 1i}, bottom). An identification accuracy of $100 \%$ was obtained with the alternative method of following 10 random animals throughout the video.

A post-processing step obtains animal images by iterative image erosion and assigns them with a heuristic (\textbf{Fig. 1j}; \textbf{Supplementary Text}). Human validation gives an accuracy of $99.988 \%$ for the final assignments, including images between crossings and during crossings.

If Protocol 2 fails, Protocol 3 starts training the convolutional part of the identification network using most of the global fragments. Then, it proceeds as Protocol 2 but always keeping the convolutional layers fixed.

We have tested idtracker.ai in small and large animal collectives (\textbf{Supplementary Tables 4} and \textbf{5}, respectively). In zebrafish, Protocol 2 was always successful, giving accuracies of $99.96$ (mean) $\pm 0.06$ (std) for $60$ individuals and $99.99$ (mean) $\pm 0.01$ (std) for 100 individuals. Importantly, of the remaining $0.01 \%$ in videos of 100 animals only $0.003 \%$ is isolated frames with assignment error and $0.007 \%$ is short non-assigned segments. In flies, Protocol 2 succeeded for a collective of $38$ individuals with $99.98\%$ accuracy. For larger groups, Protocol 3 was successful. For $72$ flies the accuracy is $99.997\%$. For $80-100$ flies the system reaches its limit, still with $>99.5\%$ accuracy.

We also studied how performance depends on the number of images between crossings.
We built synthetic global fragments obtained from individual videos of $184$ individual zebrafish (\textbf{Supplementary
Fig. 2}). We found that the system reaches high accuracy provided there is at least one global fragment with more than $30$ images per animal, but it can still be successful with fewer (\textbf{Supplementary Fig. 6}, empty markers). Recorded collectives of up to $100$ zebrafish follow this condition by a large margin (\textbf{Supplementary Fig. 6}, green dots). Flies also meet this condition except at very low locomotor activity levels here obtained in a low humidity setup (\textbf{Supplementary Fig. 6}, purple dots). Also note that conditions for video acquisition should ideally allow for a high image quality (\textbf{Supplementary Text}), but idtracker.ai seems more robust than idTracker when some of these conditions are not met (\textbf{Supplementary Table 6}).
\newline
\newline
Note: Supplementary Information is available \newline
Authors declare no conflict of interest exists

\begin{figure}[H]
\centering
\includegraphics[width=.8\textwidth]{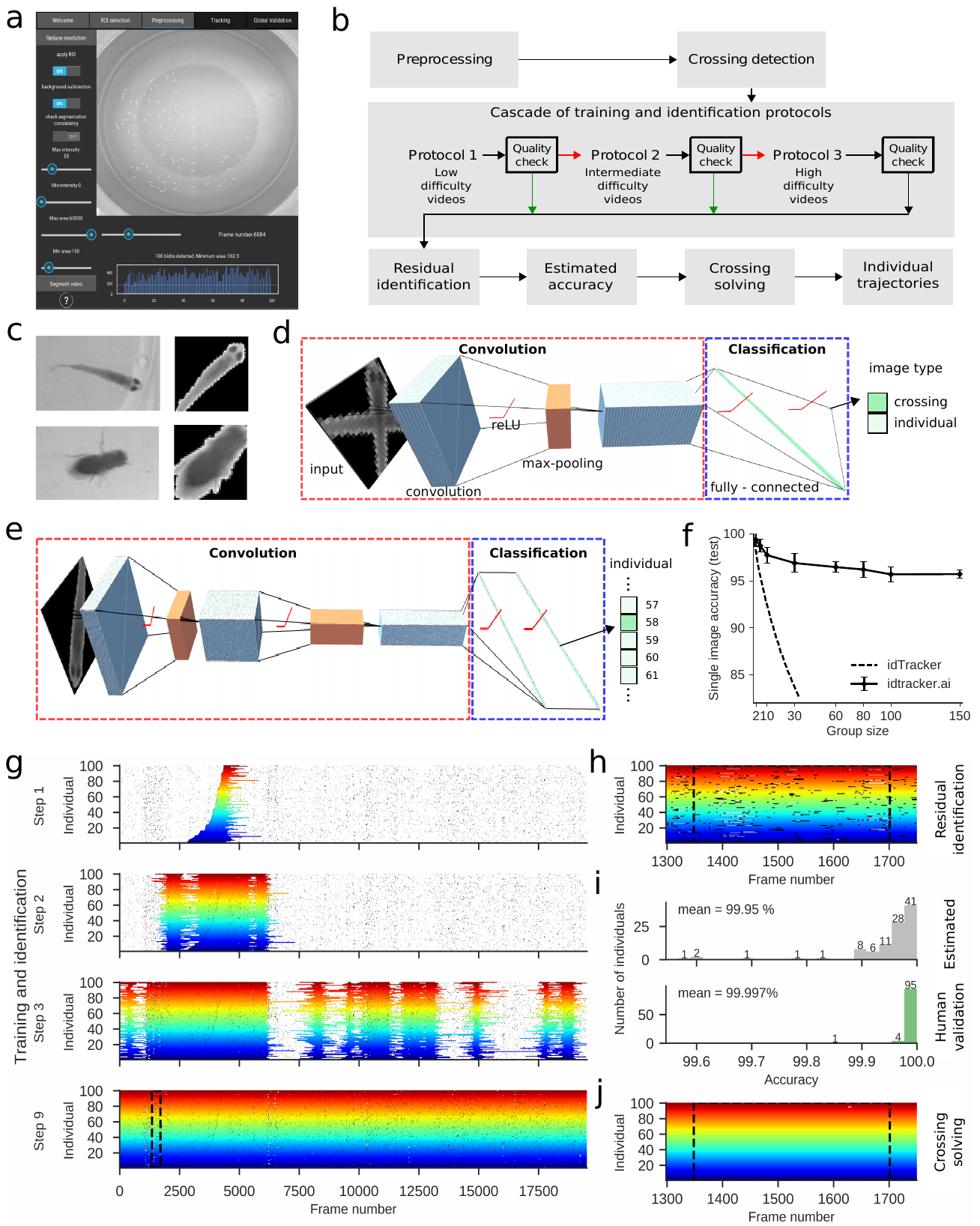}
\caption{\textbf{Tracking by identification in idtracker.ai.} \textbf{a}. Graphical user interface to track, explore and validate. \textbf{b}. Diagram of processing flow. \textbf{c}. Preprocessing to segment single animal and multi-animal blobs from background. \textbf{d}. Deep \textit{crossing detector} network is trained with examples of preprocessed single and multi-animal images. Consists of two convolutional layers (gray blocks), a max-pool operation (orange block), a hidden layer of 100 neurons (green line) and 2 output neurons to classify images into single animal/crossing. \textbf{e}. Deep \textit{identification network} is similar to d. but with 3 convolutional layers and as many output neurons as group size. \textbf{f}. Single-image accuracy as function of animal group size in original idTracker and in idtracker.ai after training with $3,000$ images per animal. \textbf{g}. Accumulation of training images in a video of 100 zebrafish of 31 dpf. Small horizontal black segments correspond to crossings detected by the \textit{crossing detector} network in d. Step 1: global fragment (portion of video in which animals do not touch or cross) with the animal that moves less having the longest distance traveled. It trains the \textit{identification network}, which in turn assigns the other global fragments, of which our quality check procedure selects a high quality subgroup (Step 2). If it does not cover enough of video, it moves to Protocol 2. Iterative training and quality checks end here in Step 9. \textbf{h}. Residual identification with final network plotted as short segments in lower transparency colors. Here shown
a zoom into the region between dashed lines in g. Small segments in white are non-assigned at this step \textbf{i}. Estimated (top) and manually validated (bottom) accuracies. \textbf{j}. Postprocessing assigns crossings and small non-assigned (white) segments in h.
}
\label{fig:fig1}
\end{figure}

\section*{Acknowledgements}
We thank Alfonso Perez-Escudero and Andres Laan for discussions, Antonia Groneberg and Andres Laan for a critical reading of the manuscript, João Bauto and Ricardo Ribeiro for assistance in hardware and software, Paulo Carri\c{c}o for help in designing
of fish arenas, Ana Catarina Certal and Isabel Campos for animal husbandry,
Andrew I. Bruce and Nico Blüthgen for videos of ants (Diacamma) and Joana Couceiro, Liliana Costa, Clara Ferreira and Tomas Cruz for assistance with fly experiments. This study was supported by a GPU NVIDIA grant (to M.G.B, F.H and G.G.dP), Funda\c{c}ao para a Ci\^{e}ncia e Tecnologia PTDC/NEU-SCC/0948/2014 (to G.G.dP.) including a contract to F.H and Champalimaud Foundation (to G.G.dP.), including contracts to M.G.B and R.H. F. R-F. acknowledges a FCT PhD fellowship.

\section*{Author contributions}

F.R-F., M.G.B. and G.G.dP. devised project, algorithms and analysed data, F.R-F. and M.G.B. wrote the code with help from F.H., M.G.B. managed code architecture and GUI, F.R-F. managed testing procedures, R.H. built set-ups and performed experiments with help from F.R-F., G.G.dP. supervised project, M.G.B. wrote supplement with help from F.R.-F, R.H, F.H. and G.G.dP., and G.G.dP. wrote main text with help from F.R.-F, M.G.B. and F.H.

%\newpage
\section*{Methods}

\subsection*{Software availability}
idtracker.ai is open-source and free software (license GPL v.3).
The source-code as well as the instructions for its installation are available in \href{https://gitlab.com/polavieja_lab/idtrackerai}{www.gitlab.com/polavieja\_lab/idtrackerai}. A quick-start user guide and a detailed explanation of the graphical user interface can be found in \href{http://idtracker.ai/}{www.idtracker.ai}.

\subsection*{Data availability}
All videos used in this study can be downloaded from \href{http://idtracker.ai/}{www.idtracker.ai}. A library of single-individual images of zebrafish to test identification methods can be found in the same link. Two example videos, one of 8 adult zebrafish and another of 100 juvenile zebrafish, are also included as part of the quick-start user guide.

\subsection*{Computers}
We tracked all the videos with desktop computers running GNU/Linux Mint 18.1 64bit (processor  Intel  Core  i7-6800K or i7-7700K, 32 or 128 GB RAM, Titan X or GTX 1080 Ti GPU's, and 1 Tb SSD disk). Sample videos can be tracked using CPU but the performance of the system will be highly affected.

\subsection*{Animal rearing and handling}
All fish were raised at the Champalimaud Foundation Fish Platform, according to the housing and husbandry methods integrated in the zebrafish welfare program fully described in \citep{martins2016toward}.
Animal handling and experimental procedures were approved by the Champalimaud Foundation Ethics Committee and the Portuguese Direcção Geral Veterinária and were performed according to the European Directive 2010/63/EU. For zebrafish videos we used the wild-type TU strain at 31 days post fertilization (dpf). Animals were kept in 8 L holding tanks at a density of 10 fish/L and a 14 h light / 10 h
dark cycle in the main fish facility. For each experiment, a holding tank with the necessary number of fish was transported to the experimental room, where fish were carefully transferred to the experimental arena using a standard fish net appropriate for their age.

\noindent For the fruit fly videos we used adults from the Canton S wild-type strain at 2-4 days post-eclosion. Animals were reared on a standard fly medium and kept on a 12-h light-dark cycle at $28^{\circ}$. Flies were placed in the arena either by anesthetizing them with CO$_2$ or ice, or by using a suction tube. We found the last method to have the least negative effect on the flies' health and to provide better activity levels.

\subsection*{Details of the networks}

\subsubsection*{Network architectures}
The deep \textit{crossing detector} network (\textbf{Fig. 1d}) is a convolutional neural network \cite{LeCun2015,Rusk2015}. It has 2 convolutional layers that obtain from data a relevant hierarchy of filters. A hidden layer of 100 neurons then transforms the convolutional output into a classification into single animal or crossing. idtracker.ai trains this network using images that can confidently characterize as single or as multiple animals (for example, single animals as blobs of area consistent with single-animal statistics and not splitting into more blobs in its past or future). Further details of the architecture are given in \textbf{Supplementary Table 1}.

The architecture of the \textit{identification} network (\textbf{Fig. 1e}) consists of 3 convolutional layers, a hidden layer of 100 neurons and a classification layer with as many classes as animals in the collective. Further details are given in \textbf{Supplementary Table 1}. We tested variations of the architecture either by modifying the number of convolutional layers \textbf{Supplementary Table 2} or the number of hidden layer neurons \textbf{Supplementary Table 3}. Analysis of these networks indicated that the most important feature for a successful identification is that the convolutional part needs at least two layers (\textbf{Supplementary Fig. 1}). The GUI allows users to modify the architecture of this network and its training hyperparameters.

\subsubsection*{Network training}

The convolutional and fully-connected layers of both networks are initialised using Xavier initialisation~\cite{glorot2010understanding}. Biases are initialised to $0$.

The deep \textit{crossing detector} network is trained using the algorithm and hyperparameters  in~\cite{kingma2014adam}. The learning rate is set at the initial value of $0.005$. This network is trained in mini batches of $100$ images.

The \textit{identification} network is trained using stochastic gradient descent, setting the learning rate to $0.005$. This network is trained in mini batches of $500$ images. Further details are given in the \textbf{Supplementary Text}.

\newpage
\begin{appendices}
	\title{Supplementary material \\ ``idtracker.ai: Tracking all individuals in large collectives of unmarked animals''}
\crefalias{section}{appsec}
\crefalias{figure}{appfig}
\crefalias{table}{apptab}
%\title{Supplementary material}
\renewcommand{\figurename}{Supplementary Figure}
\renewcommand{\tablename}{Supplementary Table}
%%%%%%%%%%%%%%%%%%%%%%%%%%%%%%%%%%%%%%%%%%%%%%%%%%%
\setcounter{figure}{0}
\setcounter{section}{0}
\maketitle
\newpage
\tableofcontents
\newpage

\section{Supplementary figures}
\begin{figure}[h!]
\centering
\includegraphics[width=\textwidth]{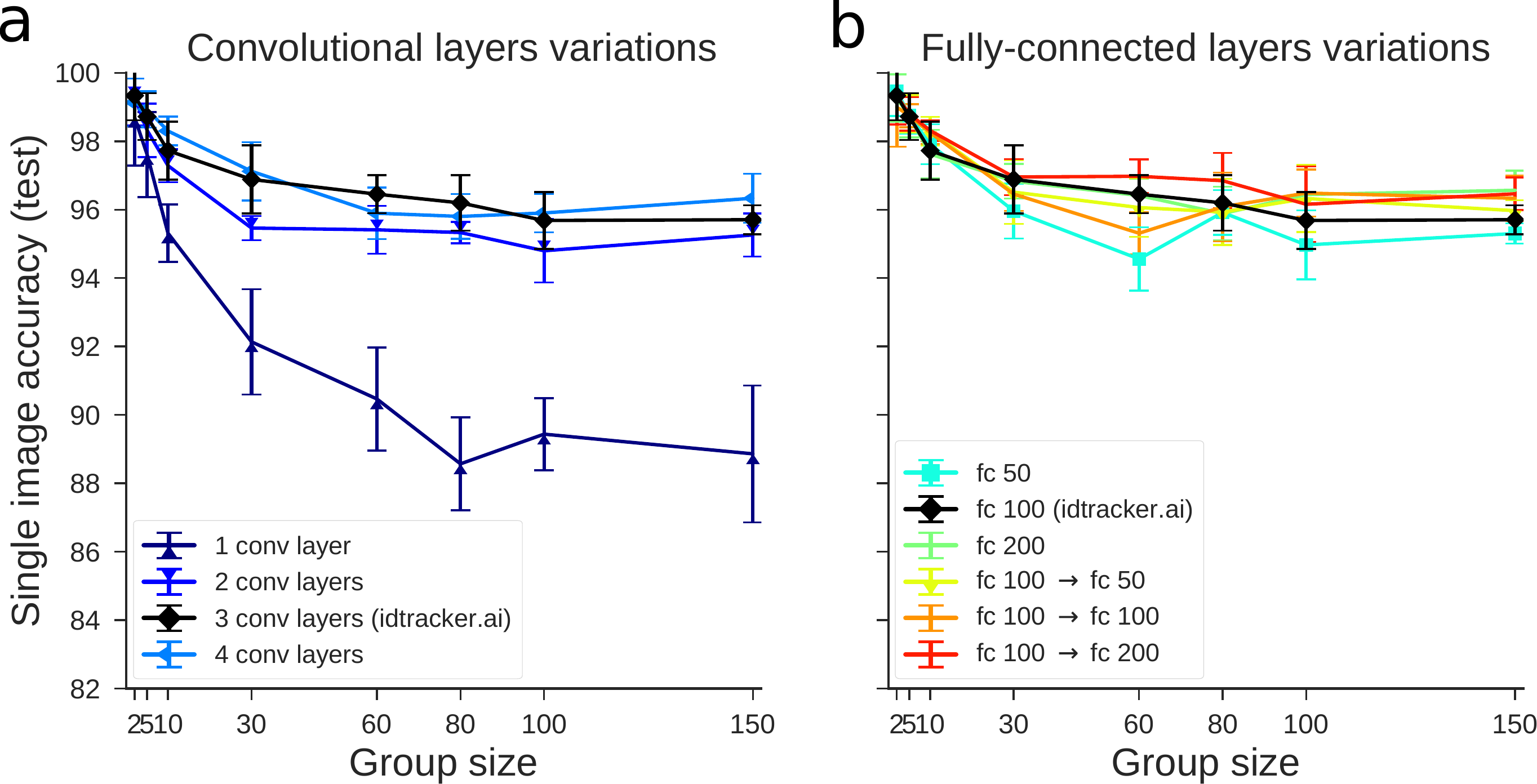}
\caption{\textbf{Single image identification accuracy for different group sizes and different variations of the identification network.} Each network is trained from scratch using $3000$ temporally uncorrelated images per animal ($90\%$ for training and $10\%$ for validation) and then tested with $300$ new temporally uncorrelated images to compute the single image identification accuracy (see~\cref{sec:identification_network}). We train and test each network five times. For every repetition the individuals of the group and the images of each individual are selected randomly. Images are extracted from videos of $184$ different animals recorded in isolation (see~\cref{fig:CARP_setup}). Colored lines with markers represent single image accuracies (mean $\pm$ std., N = 5) for networks architectures with different number of convolutional layers (\textbf{a}, see~\cref{tab:conv_architectures} for the architectures) and different size and number of fully connected layers (\textbf{b}, see~\cref{tab:conv_architectures} for the architectures). The black solid line with diamond markers shows the accuracy for the network used to identify images in idtracker.ai (see~\cref{tab:main_cnn_architectures}, Identification convolutional neural network). \label{fig:Single_image_identification_accuracy_for_different_CNN_architectures}}
\end{figure}

\newpage
\begin{figure}[h!]
\centering
\includegraphics[width=\textwidth]{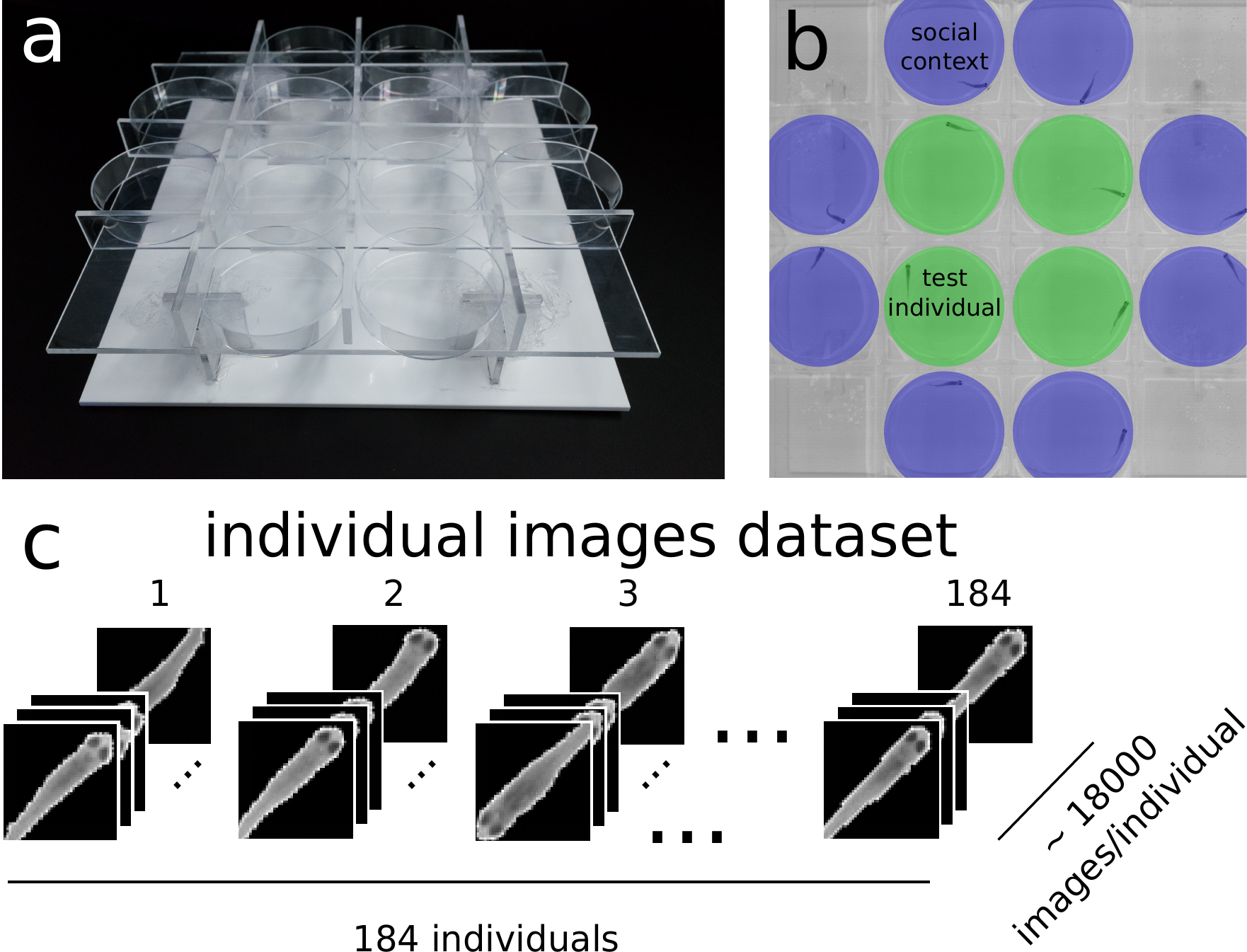}
\caption{\textbf{Creating the training dataset of individual images.} \textbf{a.} Holding grid used to record 184 juvenile zebrafish (TU strain, $31$dpf) in separated chambers ($60$ mm diameter Petri dishes). Transparent acrylic walls allowed for equal spacing between arenas, while granting visual access to the neighbouring dishes. To enhance the contrast, we used a white acrylic floor placed at a distance of $5$ cm from the holding grid, acting as a light diffuser the floor impede the formation of shadows. See~\cref{fig:fish_setup} for an explanation of the other components of the setup. \textbf{b.} Four individuals at a time were recorded for $10$ minutes (green circles). On the outer borders we placed additional dishes with fish to act as social stimuli (purple circles).
\textbf{c.} From these videos, images were labelled according to the individual they represented. Each image was preprocessed following the procedure detailed in~\cref{sec:idcnn_preprocessing}, and then cropped as a square image, in order to be used to test the identification network (image size $52\times 52$px). The dataset is composed by a total of $\approx 3312000$ uncompressed, grayscale labelled images.}\label{fig:CARP_setup}
\end{figure}

\newpage
\begin{figure}[h!]
\centering
\includegraphics[width=\textwidth]{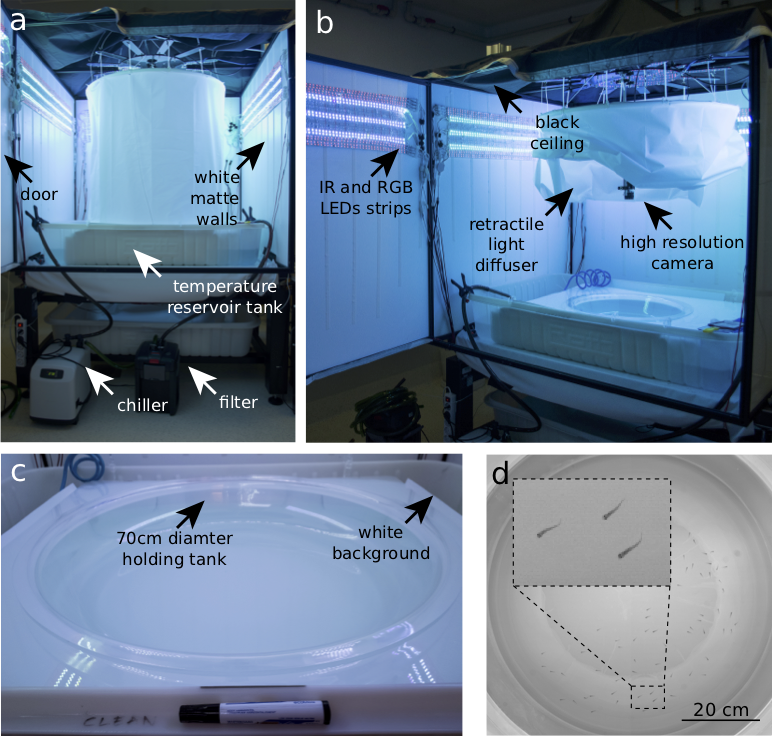}
\caption{\textbf{Experimental setup to recording zebrafish videos.} \textbf{a.} We built a setup to obtain high image quality zebrafish videos. A main tank with a water recirculating system equipped with a filter and a chiller ensures a constant water temperature of $28^{\circ}$ C. The tank is placed inside a box built with matte white acrylic walls with a door to allow for an easy access and manipulation of the setup. \textbf{b.} The lighting is based on infrared and RBG LED strips. Homogeneous illumination in the central part of the main tank is obtained by using a cylindrical retractable light diffuser made of plastic. A $20$ MP monochrome camera (Emergent Vision HT-20000M) with a $28$ mm lens (ZEISS Distagon T* 28 mm f/2.0 Lens with ZF.2) was positioned at $\approx 70$ cm from the surface of the arena. To prevent reflections of the room ceiling, a black fabric was used to cover the top of the box. \textbf{c.} We used this setup to record videos of zebrafish in groups and isolation (see~\cref{fig:CARP_setup} for details on the isolation conditions). The videos of groups of $10$, $60$ and $100$ fish were recorded in a custom-made one-piece circular tank of $70$ cm of diameter. The tank was filled up with fish system water ($28^{\circ}$ C) up to $2.5$ cm from the bottom. The circular tank was held in contact with the water of the main tank at a distance of $\approx 10$ cm from a white background to improve the contrast between the animals and the background. \textbf{d.} Sample frame from a video of $60$ animals.}
\label{fig:fish_setup}
\end{figure}

\newpage
\begin{figure}[h!]
\centering
\includegraphics[width=\textwidth]{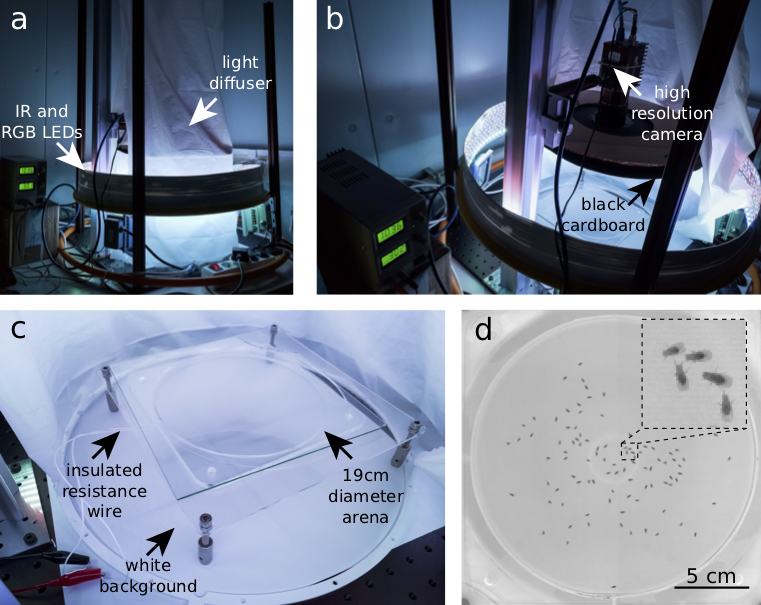}
\caption{\textbf{Experimental setup to record fruit flies videos.} \textbf{a.} The setup was placed in a dedicated experimental room with controlled humidity ($60\%$) and temperature ($25^{\circ}$ C). The illumination consisted of RBG and IR LEDs placed on a ring around a cylindrical light diffuser to guarantee homogeneous light conditions in the central part of the setup. \textbf{b.} Videos were recorded using a $20$ MP monochrome camera (Emergent Vision HT-20000M) with a $28$ mm lens (ZEISS Distagon T* 28 mm f/2.0 Lens with ZF.2) positioned at $\approx 20$ cm above the arena. Black cardboard around the camera helped to reduce reflections of the ceiling in the glass covering the arena. \textbf{c.} We used two different arenas made of transparent acrylic, both built to prevent animals from walking on the walls: Arena 1 (diameter $19$cm, height $3$mm) had vertical walls which were heated using a white insulated resistance wire (Pelican Wire Company, $28$ AWG Solid ($0.0126"$), Nichrome $60$, $4.4$ Ohms/ft, $0.015"$ White TFE Tape). At $10$ V, $0.3$ A the temperature at the walls reached $37^{\circ}$ C. Arena 2 (diameter $19$ cm, height $3.4$ mm) had conical walls (angle of inclination: $11^{\circ}$, width of conical ring: $18$ mm). Best results were obtained by recording flies from a top view as is the standard for fruit flies (see \cref{tab:larger_groups_videos}). Arena 1 was also used for bottom view recordings, where the camera was placed below the arena, pointing upward. The top of the arena consisted of a sheet of glass covered with Sigmacote SL2 (Sigma-Aldrich) which prevented the flies from walking upside down on the ceiling. A white plastic sheet was put below the arena to increase the contrast between flies and background, and the arena was separated $5$ cm from this background in order to eliminate shadows. \textbf{d.} Sample frame from a $100$ flies video. Flies were placed in the arena either by anaesthetising them with CO$_2$, ice, or by using a suction tube. We found the last method to have the least negative effect on the flies' health and provide better activity levels.\label{fig:flies_setup}}
\end{figure}

\newpage
\begin{figure}[h!]
\centering
\includegraphics[width=10cm]{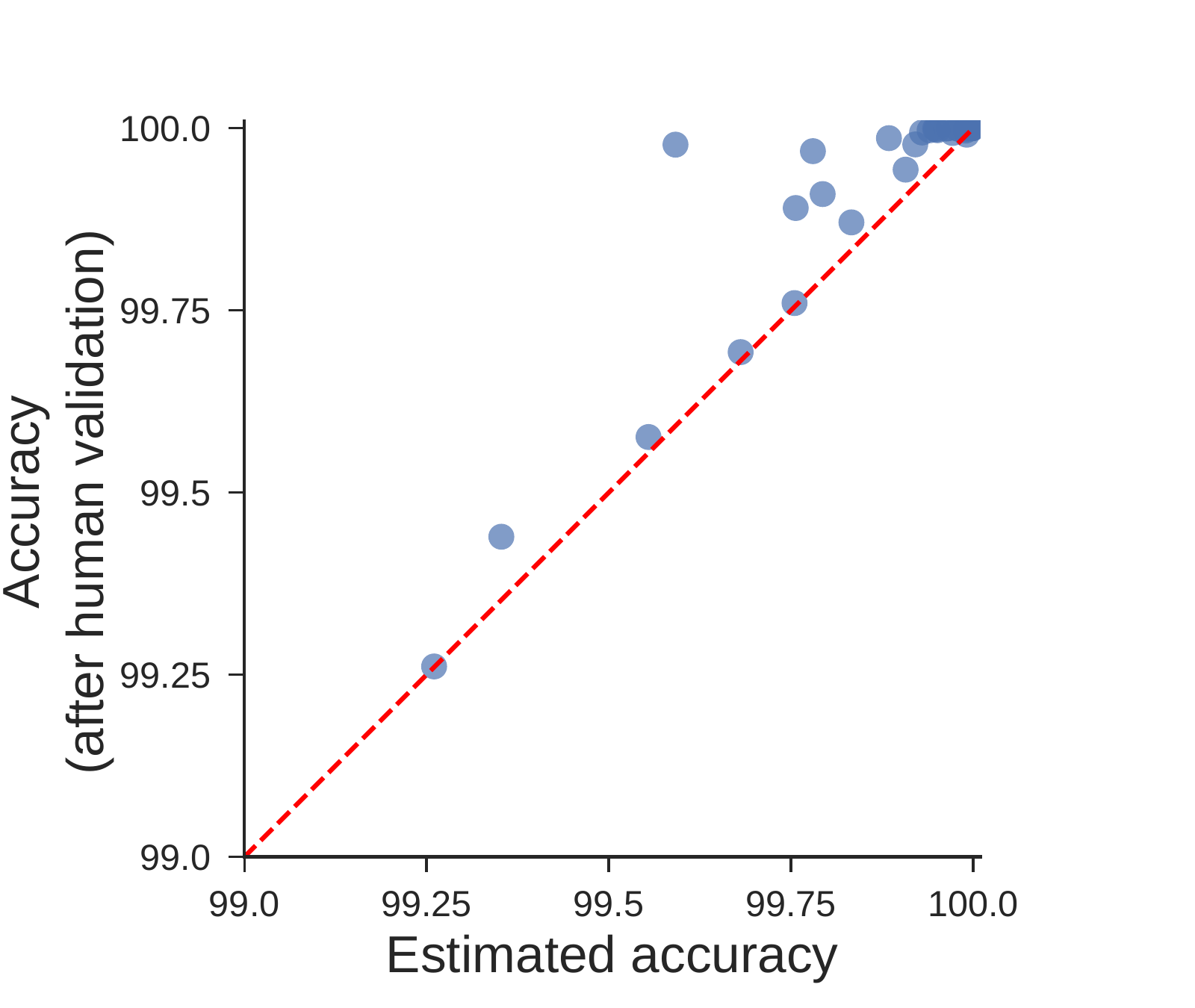}
\caption{\textbf{Automatic estimation of identification accuracy.} Comparison between the accuracy estimated automatically by idtracker.ai (see~\cref{sec:estimated_accuracy}) and the accuracy computed by human validation of the videos (see~\cref{sec:global_validation}). The estimated accuracy is computed over the validated portion of the video. Blue dots represent the videos in \cref{tab:smaller_videos}, \cref{tab:larger_groups_videos}, and \cref{tab:bad_videos_examples}. \label{fig:estimated_accuracy_vs_accuracy}}
\end{figure}

\newpage
\begin{figure}[h!]
\centering
\includegraphics[width=\textwidth]{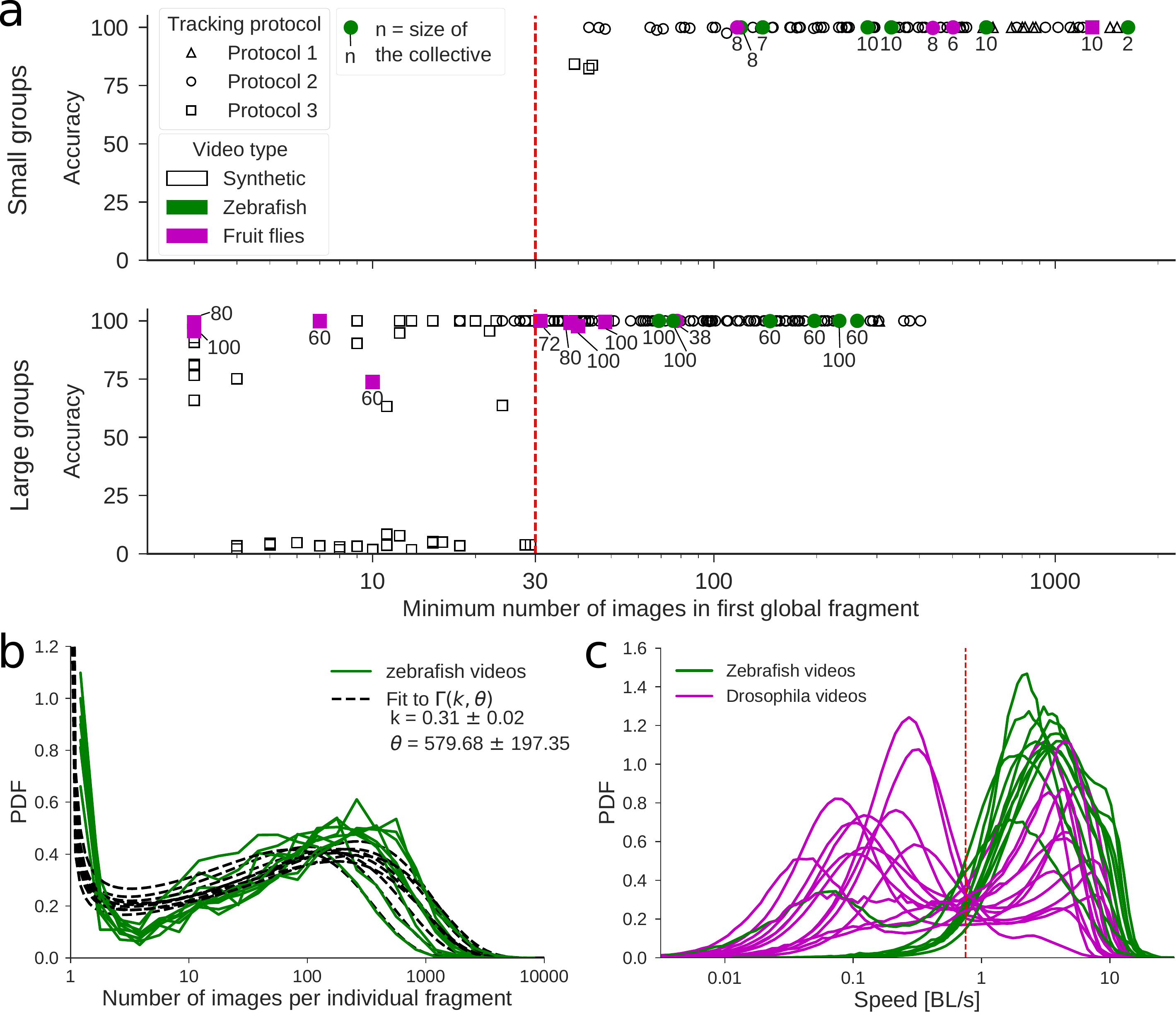}
\caption{\textbf{Accuracy as a function of the minimum number of images in the first global fragment used for training}. To study the effect of the minimum number of images per individual in the first \textit{global fragment} used to train the identification network, we created synthetic videos using images of $184$ individuals recorded in isolation (see \cref{fig:CARP_setup}). Each synthetic video consists of $10000$ frames, where the number of images in every \textit{individual fragment} was drawn from a gamma distribution and the \textit{crossings fragments} lasted for three frames (see \cref{sec:fragmentation}). The parameters were set as follows: $\theta = \left[ 2000, 1000, 500, 250, 100 \right]$, $ k = \left[0.5, 0.35, 0.25, 0.15, 0.05 \right]$, number of individual $ = \left[ 10, 60, 100 \right]$. For every combination of these parameters we ran three repetitions. In total, we computed both the \textit{training and identification protocol cascade} (see \cref{sec:deep_protocol_cascade}) and the \textit{residual identification} (see~\cref{sec:final_identification}) for $225$ synthetic videos. \textbf{a.} Identification accuracy for simulated (empty markers) and real videos (colour markers) as a function of the minimum number of images in the first global fragment. The number next to each colour markers indicates the number of animals in the video. The accuracy of the real videos is obtained by manual validation (see~\cref{tab:larger_groups_videos}, \cref{tab:smaller_videos}, and \cref{tab:bad_videos_examples}). In some videos, animals are almost immobile for long periods of time. Potentially, the individual fragments acquired during these periods encode less information useful to identify the animals. To account for this, we corrected the number of images in the individual fragments by only considering frames where the animals were moving with a speed of at least $0.75$ BL/s (body lengths per second). We observe that idtracker.ai is more likely to have higher accuracy when the minimum number of images in the first global fragment used for training is above $30$. \textbf{b.} Distributions of the number of images per individual fragments for real videos of zebrafish and their fits to a gamma distribution. \textbf{c.} Distributions of speeds of zebrafish and fruit flies videos.  \label{fig:Accuracy_vs_number_of_images_in_smaller_fragment}}
\end{figure}

\newpage
\section{Supplementary tables}
\begin{table}[h!]
\centering
\small
\caption{\textbf{Convolutional neural networks used in idtracker.ai.} The deep crossing detector network is used to classify images as belonging to a single animal or to multiple animals crossing or touching. The identification convolutional neural network (idCNN) is used to identify individual images. \label{tab:main_cnn_architectures}}
\textbf{\textit{Deep crossing detector (DCD)}} \\
\begin{tabular}{p{.1\textwidth} p{.3\textwidth} p{.2\textwidth} p{.15\textwidth} p{.1\textwidth}}
 \textbf{Layer} & \textbf{ operation} & \textbf{number of units}  & \textbf{kernel size} & \textbf{stride}\\
1                &   convolution - ReLu          & 16 & (5,5) & 1\\
\cellcolor{black!20}2                & \cellcolor{black!20}  max pooling           & \cellcolor{black!20} - &\cellcolor{black!20} - & \cellcolor{black!20}\cellcolor{black!20}(2,2)\\
3                &   convolution - ReLu          & 64 & (5,5) & 1\\
\cellcolor{black!20}4                & \cellcolor{black!20}  max pooling           &\cellcolor{black!20}  - &\cellcolor{black!20} - & \cellcolor{black!20}(2,2)\\
5                &   fully connected - ReLu           &  100 & - & -\\
\cellcolor{black!20}6                & \cellcolor{black!20}  fully connected  - softmax          &\cellcolor{black!20}  2 &\cellcolor{black!20} - &\cellcolor{black!20} -\\
\\
\end{tabular}
\small
\textbf{\textit{Identification convolutional neural network (idCNN)}} \\
\begin{tabular}{p{.1\textwidth} p{.3\textwidth} p{.2\textwidth} p{.15\textwidth} p{.1\textwidth}}
 \textbf{Layer} & \textbf{ operation} & \textbf{number of units}  & \textbf{kernel size} & \textbf{stride}\\
1                &   convolution - ReLu          & 16 & (5,5) & 1\\
\cellcolor{black!20}2                & \cellcolor{black!20}  max pooling           & \cellcolor{black!20} - &\cellcolor{black!20} - & \cellcolor{black!20}\cellcolor{black!20}(2,2)\\
3                &   convolution - ReLu          & 64 & (5,5) & 1\\
\cellcolor{black!20}4                & \cellcolor{black!20}  max pooling           &\cellcolor{black!20}  - &\cellcolor{black!20} - & \cellcolor{black!20}(2,2)\\
5                &   convolution - ReLu          & 100 & (5,5) & 1\\
\cellcolor{black!20}6                &\cellcolor{black!20}   fully connected - ReLu           &\cellcolor{black!20}  100 &\cellcolor{black!20} - &\cellcolor{black!20} -\\
7                &  fully connected  - softmax          &  group size & - & -\\
\end{tabular}
\end{table}

\begin{table}[h!]
\centering
\small
\caption{\textbf{Architectures with variations in the number of convolutional layers part}. Several architectures with variable numbers and shape of convolutional layers has been tested in order to assess the stability of the accuracy in single-image identification.\label{tab:conv_architectures}}
\begin{tabular}{c}
\textbf{\textit{1 convolutional layer}} \\
\begin{tabular}{p{.1\textwidth} p{.3\textwidth} p{.2\textwidth} p{.15\textwidth} p{.1\textwidth}}
\textbf{Layer} & \textbf{operation} & \textbf{number of units}  & \textbf{kernel size} & \textbf{stride}\\
1                &   convolution - ReLu          & 16 & (5,5) & 1\\
\cellcolor{black!20}2                &\cellcolor{black!20}   fully connected - ReLu           &\cellcolor{black!20}  100 &\cellcolor{black!20} - &\cellcolor{black!20} -\\
3                & fully connected  - softmax          &  group size & - & -\\
\end{tabular}\\
\\
\textbf{\textit{2 convolutional layers}}\\
\begin{tabular}{p{.1\textwidth} p{.3\textwidth} p{.2\textwidth} p{.15\textwidth} p{.1\textwidth}}
\textbf{Layer} & \textbf{operation} & \textbf{number of units}  & \textbf{kernel size} & \textbf{stride}\\
1                &   convolution - ReLu          & 16 & (5,5) & 1\\
\cellcolor{black!20}2                & \cellcolor{black!20}  max pooling           & \cellcolor{black!20} - &\cellcolor{black!20} - & \cellcolor{black!20}\cellcolor{black!20}(2,2)\\
3                &   convolution - ReLu          & 64 & (5,5) & 1\\
\cellcolor{black!20}4                &   \cellcolor{black!20}fully connected - ReLu &\cellcolor{black!20}  100 &\cellcolor{black!20} - &\cellcolor{black!20} -\\
5                &  fully connected  - softmax          &  group size & - & -\\
\end{tabular}\\
\\
\textbf{\textit{4 convolutional layers}}\\
\begin{tabular}{p{.1\textwidth} p{.3\textwidth} p{.2\textwidth} p{.15\textwidth} p{.1\textwidth}}
\textbf{Layer} & \textbf{operation} & \textbf{number of units}  & \textbf{kernel size} & \textbf{stride}\\
1                &   convolution - ReLu          & 16 & (5,5) & 1\\
\cellcolor{black!20}2                & \cellcolor{black!20}  max pooling           & \cellcolor{black!20} - &\cellcolor{black!20} - & \cellcolor{black!20}\cellcolor{black!20}(2,2)\\
3                &   convolution - ReLu          & 64 & (5,5) & 1\\
\cellcolor{black!20}4                & \cellcolor{black!20}  max pooling           &\cellcolor{black!20}  - &\cellcolor{black!20} - & \cellcolor{black!20}(2,2)\\
5                &   convolution - ReLu          & 100 & (5,5) & 1\\
\cellcolor{black!20}6                & \cellcolor{black!20}  max pooling           &\cellcolor{black!20}  - &\cellcolor{black!20} - & \cellcolor{black!20}(2,2)\\
7                &   convolution - ReLu          & 100 & (5,5) & 1\\
\cellcolor{black!20}8                &\cellcolor{black!20}   fully connected - ReLu           &\cellcolor{black!20}  100 &\cellcolor{black!20} - &\cellcolor{black!20} -\\
9                &  fully connected - softmax          & group size & - & -\\
\end{tabular}
\end{tabular}
\end{table}

\newpage
\begin{table}[h!]
\centering
\small
\caption{\textbf{Architectures with variations in the number and size of the fully connected layers (fc)}. Several architectures with variable numbers and shape of fully-connected layers has been tested in order to assess the stability of the accuracy in single-image identification. The notation \textit{fc $n \rightarrow$ fc $m$} characterises each architecture according to its fully-connected layers before the output layer. \label{tab:class_architectures}}
\begin{tabular}{c}
\textbf{\textit{fc 50 units}} \\
\begin{tabular}{p{.1\textwidth} p{.3\textwidth} p{.2\textwidth} p{.15\textwidth} p{.1\textwidth}}
 \textbf{Layer} & \textbf{ operation} & \textbf{number of units}  & \textbf{kernel size} & \textbf{stride}\\
1                &   convolution - ReLu          & 16 & (5,5) & 1\\
\cellcolor{black!20}2                & \cellcolor{black!20}  max pooling           & \cellcolor{black!20} - &\cellcolor{black!20} - & \cellcolor{black!20}\cellcolor{black!20}(2,2)\\
3                &   convolution - ReLu          & 64 & (5,5) & 1\\
\cellcolor{black!20}4                & \cellcolor{black!20}  max pooling           &\cellcolor{black!20}  - &\cellcolor{black!20} - & \cellcolor{black!20}(2,2)\\
5                &   convolution - ReLu          & 100 & (5,5) & 1\\
\cellcolor{black!20}6                &\cellcolor{black!20}   fully connected - ReLu           &\cellcolor{black!20}  50 &\cellcolor{black!20} - &\cellcolor{black!20} -\\
7                &  fully connected  - softmax          &  group size & - & -\\
\end{tabular}\\
\\
\textbf{\textit{fc 200 units}} \\
\begin{tabular}{p{.1\textwidth} p{.3\textwidth} p{.2\textwidth} p{.15\textwidth} p{.1\textwidth}}
 \textbf{Layer} & \textbf{ operation} & \textbf{number of units}  & \textbf{kernel size} & \textbf{stride}\\
 1-5 & (see first model) & & & \\
\cellcolor{black!20}6                &\cellcolor{black!20}   fully connected - ReLu           &\cellcolor{black!20}  200 &\cellcolor{black!20} - &\cellcolor{black!20} -\\
7                &  fully connected  - softmax          &  group size & - & -\\
\end{tabular}\\
\\
\textbf{\textit{fc 100 units $\rightarrow$ fc 50 units}} \\
\begin{tabular}{p{.1\textwidth} p{.3\textwidth} p{.2\textwidth} p{.15\textwidth} p{.1\textwidth}}
 \textbf{Layer} & \textbf{ operation} & \textbf{number of units}  & \textbf{kernel size} & \textbf{stride}\\
 1-5 & (see first model) & & & \\
\cellcolor{black!20}6                &\cellcolor{black!20}   fully connected - ReLu           &\cellcolor{black!20}  100 &\cellcolor{black!20} - &\cellcolor{black!20} -\\
7                &  fully connected  - softmax          &  50 & - & -\\
\cellcolor{black!20}8                &\cellcolor{black!20}   fully connected - ReLu           &\cellcolor{black!20}  group size &\cellcolor{black!20} - &\cellcolor{black!20} -\\
\end{tabular}\\
\\
\textbf{\textit{fc 100 units $\rightarrow$ fc 100 units}} \\
\begin{tabular}{p{.1\textwidth} p{.3\textwidth} p{.2\textwidth} p{.15\textwidth} p{.1\textwidth}}
 \textbf{Layer} & \textbf{ operation} & \textbf{number of units}  & \textbf{kernel size} & \textbf{stride}\\
 1-5 & (see first model) & & & \\\\
\cellcolor{black!20}6                &\cellcolor{black!20}   fully connected - ReLu           &\cellcolor{black!20}  100 &\cellcolor{black!20} - &\cellcolor{black!20} -\\
7                &  fully connected  - softmax          &  100 & - & -\\
\cellcolor{black!20}8                &\cellcolor{black!20}   fully connected - ReLu           &\cellcolor{black!20}  group size &\cellcolor{black!20} - &\cellcolor{black!20} -\\
\end{tabular}\\
\\
\textbf{\textit{fc 100 units $\rightarrow$ fc 200 units}} \\
\begin{tabular}{p{.1\textwidth} p{.3\textwidth} p{.2\textwidth} p{.15\textwidth} p{.1\textwidth}}
 \textbf{Layer} & \textbf{ operation} & \textbf{number of units}  & \textbf{kernel size} & \textbf{stride}\\
 1-5 & (see first model) & & & \\
\cellcolor{black!20}6                &\cellcolor{black!20}   fully connected - ReLu           &\cellcolor{black!20}  100 &\cellcolor{black!20} - &\cellcolor{black!20} -\\
7                &  fully connected  - softmax          &  200 & - & -\\
\cellcolor{black!20}8                &\cellcolor{black!20}   fully connected - ReLu           &\cellcolor{black!20}  group size &\cellcolor{black!20} - &\cellcolor{black!20} -\\
\end{tabular}
\end{tabular}
\end{table}

\newpage
\begin{landscape}
\begin{center}
\small
\begin{tabular}{p{45mm} p{13mm} p{15mm} p{15mm} p{13mm} p{15mm} p{20mm} p{23mm} p{15mm} p{15mm}}
\toprule
  Video & Duration & Frames per second & Pixels per animal & Protocol & Reviewed crossings & Accuracy prot. cascade & \textbf{Accuracy} & $\%$ Non-identified & $\%$ Mis-identified \\
\midrule
      2 nacre zebrafish (1) $\dagger$& 15'58" &               24 &               764 &        2 &                           90 &                      100 &                 \textbf{99.991}  &                0 &               0.009 \\
  7 inbred WIK zebrafish $\dagger$& 08'19" &               31 &               670 &        2 &                          393 &                      100 &                 \textbf{99.997}  &                0.003 &               0 \\
          8 zebrafish $\dagger$ $\ast$& 05'22" &               28 &               331 &        2 &                          842 &                      100 &                 \textbf{99.969}  &                0.028 &               0.003 \\
          10 zebrafish & 10'03" &               32 &               528 &        2 &                          311 &                      100 &                    \textbf{100} &                0 &               0 \\
          10 zebrafish & 10'03" &               32 &               551 &        2 &                          877 &                      100 &                 \textbf{99.999} &                0 &               0.001 \\
          10 zebrafish & 10'10" &               32 &               534 &        2 &                          667 &                      100 &                    \textbf{100} &                0 &               0 \\
\midrule
           5 medaka fish $\dagger$& 21'41" &               28 &               232 &        2 &                          124 &                      100 &                 \textbf{   100}  &                0 &               0 \\
          10 medaka fish $\dagger$& 12'20" &               28 &               199 &        2 &                           89 &                      100 &                 \textbf{   100}  &                0 &               0 \\
		20 medaka fish $\dagger$& 08'20" &               29 &               273 &        2 &                          597 &                      100 &                 \textbf{99.999}  &                0.001 &               0 \\
\midrule
       6 Drosophila (4 fem, 2 mal) $\dagger$& 07'57" &               42 &              1048 &        2 &                          306 &                      100 &                 \textbf{99.994}  &                0.006 &               0 \\
       8 female Drosophila (1) $\dagger$& 03'07" &               28 &               819 &        2 &                          199 &                      100 &                 \textbf{99.986}  &                0 &               0.014 \\
		8 female Drosophila (2) $\dagger$& 14'58" &               28 &               750 &        2 &                          298 &                   99.882 &                 \textbf{99.760}  &                0.104 &               0.136 \\

\midrule
      2 agouti mice $\dagger$& 05'19" &               49 &             12855 &        2 &                           93 &                   99.932 &                 \textbf{99.910}  &                0 &               0.090 \\
       2 black mice (1) $\dagger$& 12'24" &               49 &             11986 &        2 &                          200 &                   99.582 &                 \textbf{99.576}  &                0 &               0.424 \\
       2 black mice (2) $\dagger$& 07'06" &               49 &             11401 &        2 &                           41 &                   99.738 &                 \textbf{99.693}  &                0 &               0.307 \\
       2 black mice (3) $\dagger$& 07'06" &               49 &             12124 &        2 &                           44 &                      100 &                 \textbf{   100}  &                0 &               0 \\
       4 black mice (2)$\dagger$& 33'58" &               25 &              7398 &        2 &                          351 &                      100 &                 \textbf{99.993}  &                0.002 &               0.005 \\
       4 black mice (3)$\dagger$& 52'54" &               24 &              6440 &        2 &                          287 &                   97.974 &                 \textbf{96.641}  &                0.032 &               3.327 \\

 \bottomrule
\end{tabular}
\small
\captionof{table}{\textbf{Results of manual validation for video of small group size}. To compare the performance of idtracker.ai with respect to idTracker, we tracked and manually validated most of the videos used in~\cite{Perez-Escudero2014}. We also add three more videos of 10 zebrafish (TU strain, 31 dpf). We observe that the performance is comparable to the one obtained by idTracker (see Supplementary Table 1 in~\cite{Perez-Escudero2014}). The column ``Reviewed crossings'' displays the number of \textit{crossing fragments} as defined in~\cref{sec:fragmentation} in the validated part. The column ``Accuracy prot. cascade'' displays the proportion of individual images correctly identified after the protocol cascade. The column ``Accuracy'' displays the proportion of individual images correctly identified in the validated part. The column ``Non-identified'' displays the percentage of individual images for which the system did not assign an identity. The column ``Misidentified'' displays the percentage of individual images wrongly identified.\label{tab:smaller_videos}}
\end{center}
\small
\textbf{Table legend:} $\ast$: used to develop the tracking system. $\dagger$ video from \cite{Perez-Escudero2014}.
\end{landscape}

\newpage
\begin{landscape}
\begin{center}
\small
\begin{tabular}{p{35mm} p{13mm} p{15mm} p{15mm} p{13mm} p{15mm} p{20mm} p{23mm} p{15mm} p{15mm}}
\toprule
  Video & Duration & Frames per second & Pixels per animal & Protocol & Reviewed crossings & Accuracy prot. cascade & \textbf{Accuracy} \;\;\;\;\;\;\;\;/ Indiv. acc. & $\%$ Non-identified & $\%$ Mis-identified \\
\midrule
60 zebrafish & 10'29" &               31 &               325 &        2 &                          338 &                   99.921 &                 \textbf{99.871} / 99.880 &               0.024 &               0.106 \\
60 zebrafish & 10'30" &               31 &               308 &        2 &                          277 &                      100 &                 \textbf{99.998} / 99.999 &                0.002 &               0 \\
60 zebrafish & 10'09" &               32 &               303 &        2 &                          213 &                      100 &                    \textbf{100} / 99.999 &                0 &               0 \\
100 zebrafish & 10'16" &               32 &               316 &        2 &                         2552 &                      100 &                 \textbf{99.977} / 99.989 &              0.016 &               0.007 \\
100 zebrafish & 10'12" &               32 &               273 &        2 &                          750 &                      100 &                 \textbf{99.999} / 99.999 &                0.001 &               0 \\
100 zebrafish $\ast$ & 10'10" &               32 &               309 &        2 &                          628 &                      100 &                 \textbf{99.997} / 100 &                0.002 &               0.001 \\
\midrule
 38 fruit flies $\downarrow$ (2) & 10'07" &               36 &               602 &        2 &                          753 &                   99.996 &                 \textbf{99.978} / - &                0.002 &               0.020 \\
  60 fruit flies \Female $\uparrow$ (1)& 10'00" &               51 &               372 &        3 &                          699 &                      100 &                 \textbf{99.999} / - &          0.001 &               0 \\
  72 fruit flies $\downarrow$ (2)& 10'00" &               36 &               602 &        3 &                          165 &                      100 &                 \textbf{99.997} / - &                0.003 &               0 \\
  80 fruit flies $\downarrow$ (1)& 10'02" &               34 &               589 &        3 &                          514 &                   99.925 &                 \textbf{99.439} / - &                0.420 &               0.141 \\
  80 fruit flies $\uparrow$ (2) & 10'02" &               51 &               273 &        3 &                          222 &                      100 &                 \textbf{99.891} / 99.046 &            0.109 &               0 \\
  100 fruit flies $\downarrow$ (1) & 10'00" &               28 &               558 &        3 &                          107 &                   99.909 &                 \textbf{99.580} / - &                0.327 &               0.093 \\
\bottomrule
\end{tabular}
\captionof{table}{\textbf{Results of manual validation for videos of large group size}. To validate the system we manually reviewed the identities of the animals before and after every crossing image and every non-identified image for a part of the video (see Global Validation in \cref{sec:global_validation}). For some videos we also reviewed the entire video for 10 randomly chosen animals (see Individual Validation in \cref{sec:individual_validation}). ``Reviewed crossings'', ``Accuracy'', ``Accuracy prot. cascade'', ``Non-identified'' and ``Misidentified'' refer to the global validation, and ``Indiv. acc.''' refers to the individual validation. The column ``Reviewed crossings'' displays the number of \textit{crossing fragments} as defined in~\cref{sec:fragmentation} in the validated part. The column ``Accuracy prot. cascade'' displays the proportion of individual images correctly identified (PIICI) after the protocol cascade. The column ``Accuracy'' displays the PIICI in the validated part. The column ``Indiv. acc.'' displays the average PIICI for the 10 validated individuals. The column ``Non-identified'' displays the percentage of individual images for which the system did not give an identity. The column ``Misidentified'' displays the percentage of individual images wrongly identified. All zebrafish videos were recorded in the setup described in~\cref{fig:fish_setup}. All the fruit flies videos were recorded in the setup described in~\cref{fig:flies_setup}. \label{tab:larger_groups_videos}}
\end{center}
\small
\textbf{Table legend:} $\ast$: used to develop the tracking system and video in \textbf{Fig 1.} in main text. (1)/(2): video recorded in arena 1 or 2 respectively (see~\cref{fig:flies_setup}). $\uparrow$ or $\downarrow$: recorded from below ($\uparrow$) or from the top ($\downarrow$) respectively. \Female: the video realised with only female fruit flies.
\end{landscape}

\newpage
\begin{landscape}
\begin{center}
\small
\begin{tabular}{p{35mm} p{13mm} p{15mm} p{15mm} p{13mm} p{15mm} p{20mm} p{23mm} p{15mm} p{15mm}}
\toprule
  Video & Duration & Frames per second & Pixels per animal & Protocol & Reviewed crossings & Accuracy prot. cascade & \textbf{Accuracy} & $\%$ Non-identified & $\%$ Mis-identified \\
\midrule
\\
\end{tabular}
\\
\textbf{Compressed videos with bad illumination conditions (users reported difficulties using idTracker \cite{Perez-Escudero2014})}
\begin{tabular}{p{35mm} p{13mm} p{15mm} p{15mm} p{13mm} p{15mm} p{20mm} p{23mm} p{15mm} p{15mm}}
  10 fruit flies & 10'12" &               60 &               420 &        3 &                          288 &                    100 &               \textbf{100}  &                0 &               0 \\
        14 ants & 13'10" &               59 &               573 &        2 &                          498 &                   99.989 &           \textbf{99.943} &                0.001 &               0.056 \\
\\
\end{tabular}
\\
\textbf{Very low locomotor activity levels for most animals or dead animals}
\begin{tabular}{p{35mm} p{13mm} p{15mm} p{15mm} p{13mm} p{15mm} p{20mm} p{23mm} p{15mm} p{15mm}}
  60 fruit flies & 10'14" &               51 &               292 &        3 &                           - &                 - &                 73.720 $\dagger$  &             - &            - \\
   100 fruit flies \Female & 10'14" &               50 &               334 &        3 &                           - &                 - &                 95.579 $\dagger$  &             - &            - \\
\\
\end{tabular}
\\
\textbf{Some animals show atypical behavior with poses very different to healthy ones}
\begin{tabular}{p{35mm} p{13mm} p{15mm} p{15mm} p{13mm} p{15mm} p{20mm} p{23mm} p{15mm} p{15mm}}
100 fruit flies (2) & 10'00" &               34 &               538 &        3 &                          217 &                   99.905 &          \textbf{99.277}  &                0.546 &               0.177 \\
\\
\bottomrule
\end{tabular}
\small
\captionof{table}{\textbf{Results of manual validation for videos that do not fulfil some of the general video conditions listed in \cref{sec:video_conditions}}. The column headers are defined in~\cref{tab:smaller_videos}. The first two videos were recorded in a compressed lossy video formats, .avi (FMP4 compression code) and .MOV (avc1), respectively. Lossy video formats may deleted pieces of information that could be important to identify the animals. The video of 10 fruit flies was recorded in a retroilluminated circular arena with conic walls. During the video, at regular time intervals a LED turns on and off. In images from retroilluminated arenas animals appear mainly as dark blobs, and the visual features of the body of the animals are less visible resulting in a more difficult identification. In addition, changes in light conditions can be problematic, since the appearance of the animals can change. This video was recorded by Clara Ferreira. The video of $14$ ants has multiple shadows of animals which make the segmentation difficult. Furthermore, it is not illuminated with indirect light creating reflections in the body of the animals. This videos was given by Andrew. I. Bruce and Nico Blüthgen (Diacamma ants, preliminary tracking tests). The animals in the video of $60$ flies have very low locomotor activity levels, and the first video of $100$ female flies contains dead animals due to the after effects of the anaesthesia with CO$_2$. Images of animals during immobility periods represent a small set of the poses that animals can show over the duration of the video. A network trained with images of immobile animals is likely to fail to identify an animal when it starts moving or changes its pose. Moreover, the video of $60$ flies has poor contrast and was segmented with a bad selection of preprocessing parameters. In the second video of $100$ flies some animals show atypical behaviours like rolling on their back. In particular, there is an animal rolling on its back in the first global fragment. After with this global fragment, other animals which roll on their backs are identified as this animal.\label{tab:bad_videos_examples}}
\end{center}
\small
\textbf{Table legend:} $\dagger$: Estimated accuracy (see~\cref{sec:estimated_accuracy}).
\end{landscape}

\newpage
\section{General video conditions \label{sec:video_conditions}}

It is advisable to adhere to some guidelines during the realisation of videos of freely-moving animals. Here follows a list of conditions that allow to maximise the probability of success and the accuracy of the tracking.

\begin{itemize}

\item \textbf{Resolution}. The higher the number of pixels per individual, the more information to distinguish it from the rest. Notice that, on the downside, the additional information makes the algorithm less time-efficient. Check \cref{tab:larger_groups_videos,tab:smaller_videos} for the average number of pixels per animal in each of the videos tracked.

\item \textbf{Frame rate}. The frame rate must be high enough for the blobs associated with the same individual to overlap in consecutive frames, when moving at average speed. A low frame rate---with respect to the average speed of the animals---can cause a bad fragmentation of the video: An essential process in the tracking pipeline, that allows to collect images belonging to the same individual and organise them in fragments. On the contrary, excessively high frame rates will make the information coming from the analysis of the fragments highly redundant. This will increase the computational time necessary to track the video, without guaranteeing an improvement of the identification of the individuals. In the examples provided in this paper, the frame rate ranges from $25$fps to $50$fps.

\item \textbf{Duration}. The length of the video for which the system works depends on the number of animals, the distribution of images per individual fragment and the number of pixels per animal. For few animals (8 zebrafish) we can track videos as short as $\approx 18$ sec ($\approx 500$ frames at 28 fps. For large groups we can track videos as short as $1$ min ($\approx 1950$ frames at $32$ fps). The system works for longer videos as far as the overall conditions do not change abruptly in different parts of the video. Very large  videos with many animals will require a high amount of RAM and could block your computer.

\item \textbf{Video format}. The system works with any video format compatible with OpenCV. We recommend uncompressed or lossless video formats: Some compression algorithms work by deleting pieces of information that could be crucial for the identification of the individuals. However, we have successfully tracked videos with compressed formats: .avi (FPM4 video codec) and .MOV (avc1 video codec) (see~\cref{tab:bad_videos_examples}).

\item \textbf{Illumination}. Illumination has to be as uniform as possible, so that the appearance of the animals is homogeneous along the video. We recommend using indirect light either by making the light reflect on the walls of the setup, or by covering the setup with a light diffuser as shown in~\cref{fig:fish_setup,fig:flies_setup}. Although, we have also tracked videos with retroilluminated arenas (see~\cref{tab:bad_videos_examples}), recall that the tracking systems relies on visual features of the animals that this type of illumination could hide.

\item \textbf{Definition and focus}. Images of individuals should be as sharp and focused as possible for their features to be clearly displayed along the entire video. When using wide apertures on the camera, the depth of field can be quite narrow. Make sure that the plane of the sensor of the camera is parallel to the plane of the arena so that animals are focused in all parts of it. In addition, exposition time (shutter speed) should be high enough so that animals do not appear blurred when moving at average speed. Blurred and out of focus images are more difficult to be identified correctly.

\item \textbf{Background}. The background should be as uniform as possible. To facilitate the detection of the animals during the segmentation process (see~\cref{sec:segmentation}), the background colour has to be chosen in order to maximise the contrast with the animals. Small background inhomogeneity or noise are acceptable and can be removed by the user during the selection of the preprocessing parameters:
\begin{itemize}
\item Static or moving objects much smaller or much larger than the animals can be removed by setting the appropriate maximum and minimum pixels size thresholds.

\item Static objects of the same size and intensity of the animals can be removed by selecting the option “subtract background” in the preprocessing tab.

\item Regions of the frame can be also excluded by selecting a region of interest.
\end{itemize}

\item \textbf{Shadows}. Shadows projected by the individuals on the background can lead to a bad segmentation and hence, to a bad identification. Shadows can be diffused by using a transparent base separated from an opaque background (see~\cref{fig:fish_setup}) or by using a retroilluminated arena.

\item \textbf{Reflections}. Reflections of individuals on the walls of the arena should be avoided: They could be mistaken for an actual individual during the segmentation process. Reflections in opaque walls can be reduced by using either very diffused light or matte walls. For aquatic arenas with transparent walls, reflections can be softened by having water at both sides of the walls. Furthermore, reflections can be removed by selecting an appropriate ROI.

\item \textbf{Variability in number of pixels per animal}. The number of pixels in a blob is one of the criteria used to distinguish individual fish from crossings. An optimal video should fulfil the two following conditions. First, the number of pixels associated with each individual should vary as little as possible along the video. Second, the size an individual should vary as little as possible depending on its position in the arena. In any case, strategies to avoid misidentification are put in place, even in case of variable animal sizes (see~\cref{sec:crossing_detector}).
\end{itemize}

\section{Algorithm\label{sec:algorithm}}

\begin{figure}[tb]
\centering
\includegraphics[width=.5\textwidth]{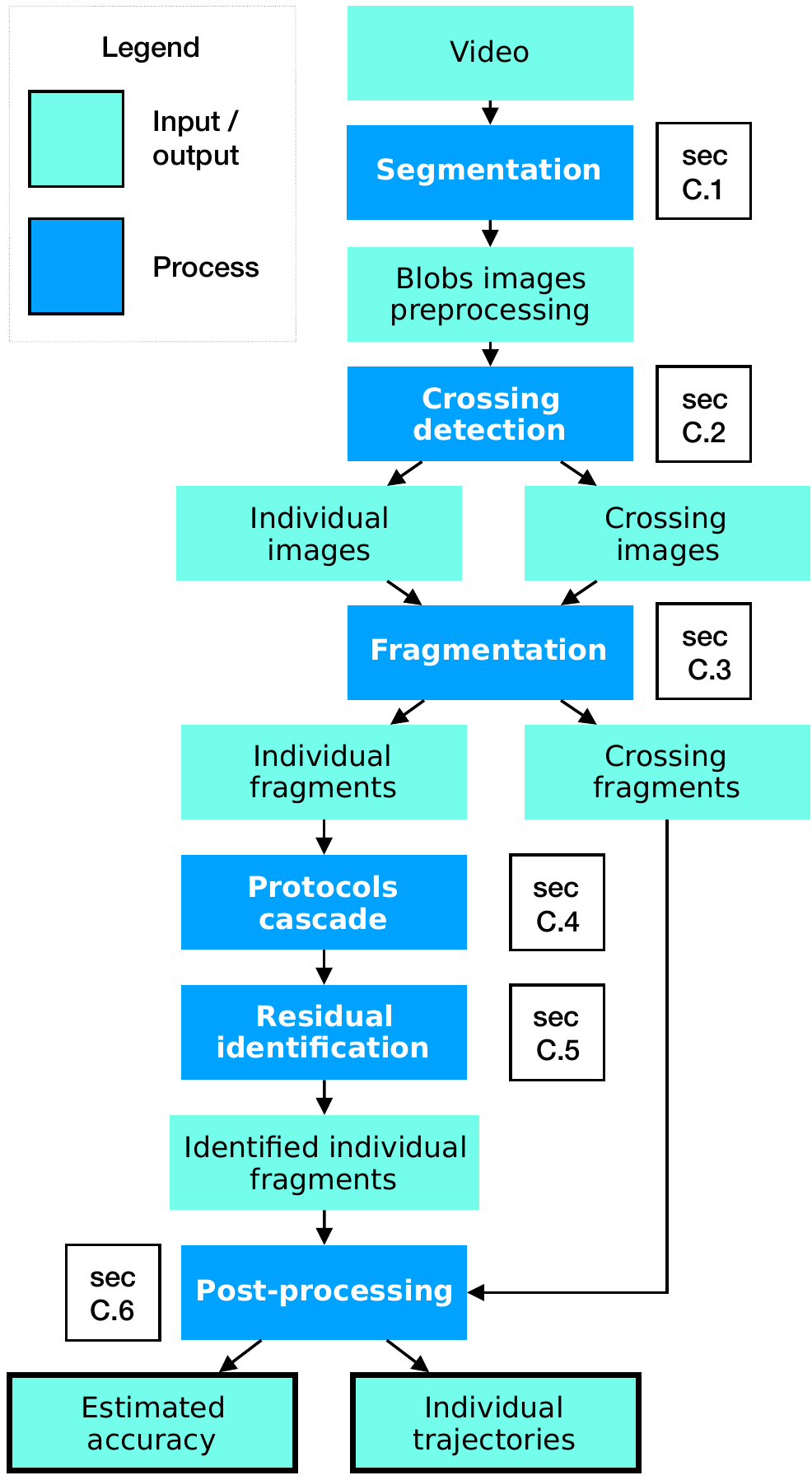}
\caption{Simplified algorithmic flow. Refer to the section specified next to each process block for details.\label{fig:alg_flow}}
\end{figure}

First, we introduce the work-flow of the algorithm. Subsequent sections will give further details on each of the components in the work-flow. The algorithm is composed of six computational cores highlighted in blue in~\cref{fig:alg_flow}. First, during the segmentation process the images representing either single or multiple touching animals are extracted from the video. In the remainder, we will refer to images representing a single individual as \textit{individual images} and to images in which two or more individuals are touching as \textit{crossing images}.

A model of the average area of the individuals, and later a convolutional neural network (CNN)---named deep crossing detector in the remainder---are used to discriminate between individual and crossing images.

Each image extracted from the video is now labelled as either a single individual or a crossing.
By means of an extra-safe protocol, we define collections of images in subsequent frames of the video in which the same individual (or crossing) is represented. We name these collections individual and crossing \textit{fragments}, respectively.

The fourth computational core is the gist of the algorithm. A subset of the collection of individual fragments, in which all the individuals are visible in the same part of the video is used to generate a dataset of individual images labelled with the corresponding identities. This dataset is then utilised to train a second CNN to classify images according to their identity. A cascade of increasingly encompassing training/identification protocols is put in place, so that an appropriate identification strategy is automatically defined by the algorithm, according to the complexity of the video. The idea underlying this family of methods is that the information gained from the first dataset of labelled images will allow either to accurately assign the entire collection of individual fragments, or to increase the first dataset by incorporating safely identified individual fragments throughout the video.

The knowledge acquired during the protocol cascade is used to identify the individual fragments that were not used to train the identification CNN. In the remainder, we will refer to this operation as \textit{residual identification}.

Finally, trivial identification errors are corrected by a series of post-processing routines, and the identity of the crossing fragments is inferred in a last computational core.
%%%%
% Segmentation
%%%%
\subsection{Segmentation\label{sec:segmentation}}

\begin{figure}[h]
\centering
\includegraphics[width=.6\textwidth]{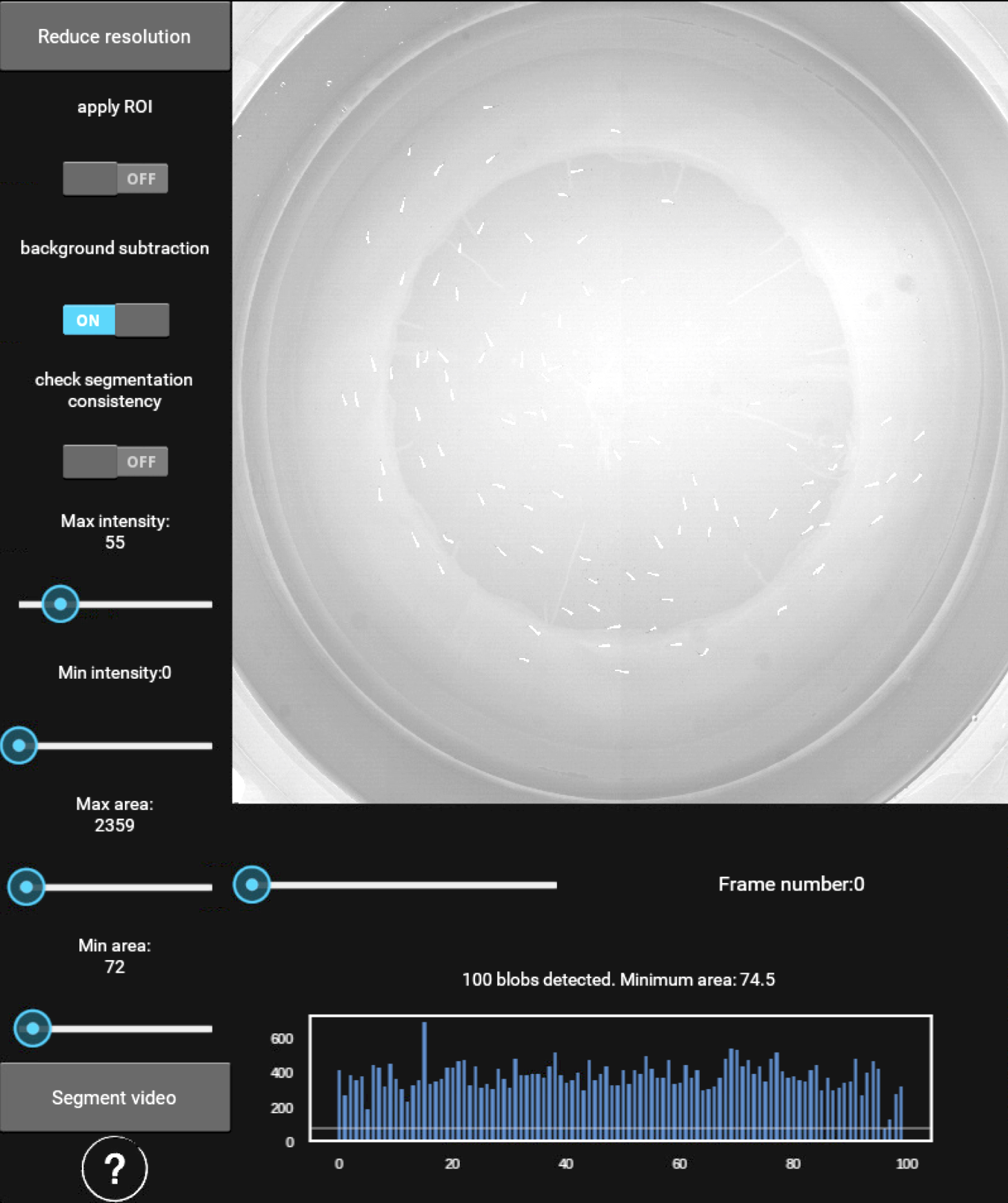}
\caption{Grayscale image thresholding for blob segmentation. idtrackerai has been tested with average individual blob areas of $\sim 300$pxs. The resolution reduction button (top-left) allows to introduce a downsampling factor, to be applied to the entire frame, and consequently to the segmented blobs. The tree switches on the top allow to consider a predefined ROI, compute and subtract a model of the background. It is possible to activate a control on the number of blobs detected in each frame: If the segmentation returns more blobs than animals to be tracked in a frame or a collection of frames, the user will be asked to specify new segmentation parameters to be applied only in those frames. Finally, the user can define ranges of both acceptable intensities and blob areas, by adjusting the \textit{Maximum} and \textit{Minimum} intensity and area thresholds respectively. } \label{fig:thresholding}
\end{figure}

idtracker.ai tracks the individuals by relying on their visual features. Hence, given a frame of the video, it is necessary to distinguish between pixels associated to individuals and background. According to the standard notation adopted in computer vision, we refer to a collection of connected pixels which is not part of the background as a \textit{blob}.

The segmentation process has four main steps. First, the user can define a region of interest to be applied on each frame of the video. In this way it is possible to exclude, for instance, walls which may contain reflections of the animals.

Second, each frame is normalised with respect to its average intensity to correct for illumination fluctuations. It is also possible to perform background subtraction by generating a model of the background calculated as the average of a collection of frames obtained via subsampling the video.

Then, blobs of pixels corresponding to animals are detected by intensity thresholding and subsequent labelling of connected components. The intensity thresholds that allow to distinguish the individuals from the background are specified by the user. Often, intensity is not enough to segment the animals in the entire video. For this reason, it is also possible to specify a minimum and a maximum area (number of pixels) for a blob to be acceptable. For instance, these parameters allow to exclude dust during the segmentation.

All these operations are carried out in an intuitive way by using the idtracker.ai graphic user interface, where both the intensity and area thresholds can be adjusted by observing their effect in real time on the frame, see~\cref{fig:thresholding}.

The software currently supports only grayscale video segmentation. Frames captured from a color video will be automatically mapped to grayscale.

\begin{rem}[On background subtraction]
Background subtraction is often useful when trying to segment a video in which a static object has the same intensity level as the individuals one wants to segment (see \cref{sec:video_conditions}).
\end{rem}

\subsection{Detection of individual and crossing images\label{sec:crossing_detection_CNN}}

The training/identification process allows to identify only images representing single individuals. Thus, a crucial point of the algorithm is the discrimination of individual and crossing images. In order to differentiate between these two classes, we apply a series of three different algorithms on the images segmented from the video.

First, we use two heuristics to detect images that in all likelihood correspond to a single animal (sure individual image) and crossing animals (sure crossing images), respectively. Then, we use these sure individual images and sure crossing images to train a neural network.
Finally, the trained network is used to label ambiguous (not sure) images as either crossing or individual images.

\subsubsection{Model area\label{sec:model_area}}
We build a model of the area of the individuals by taking into account portions of the video in which the number of segmented blobs corresponds to the number of animals declared by the user. In case there is no frame in which this condition is fulfilled, the tracking cannot proceed and an error is raised. Let $\mathcal{C} = \left\{ b_1, \dots, b_n\right\}$ be the collection of the blobs segmented from these parts of the video and $A = \left\{area\left(b_i \right) \mbox{ for every } b_i\in \mathcal{C} \right\}$ the collections of the corresponding individual areas, where the function $area(b_i)$ counts the number of pixels corresponding to the blob $b_i$. The model area is defined by $m_A = \mbox{median}\left(A \right)$ and the standard deviation $s_A = \sigma\left(A\right)$. Let $b$ be a blob, we define
\begin{equation}\label{eq:model_area}
\gamma\left(b \right) = \begin{cases}
\mbox{is an individual} & \mbox{if }  \left| area\left( b \right) - m_A \right| < 4 \cdot s_A\\
\mbox{is a crossing} & \mbox{ otherwise}
\end{cases}
\end{equation}

A model based exclusively on the area of the blobs can easily fail when the individuals' body is not rigid (e.g.~fish or mice), can suddenly change shape (e.g.~a fly with opened or closed wings), or under heterogeneous lighting conditions. Even more complex situations can arise when animals can move freely in 3 dimensions (\eg fish swimming at different depths). In this latter case, one individual can be almost completely occluded by a second one, causing the model area to fail.

\subsubsection{Blobs overlapping in subsequent frames\label{sec:segmented_images_overlapping}}
The second heuristic is based on the overlapping of blobs in subsequent frames: This allows to select sure crossing and individual images depending on the merging or splitting of consecutive, overlapping blobs.
We recall that a blob is a collection of connected acceptable pixels in a certain frame, where a pixel is considered acceptable depending on its intensity value and the thresholding described in~\cref{sec:segmentation}. Let $b_1$ and $b_2$ be two blobs. We say that the two blobs overlap if and only if $b_{1}\cap b_{2} \neq \emptyset$, where the intersection $b_{1}\cap b_{2}$ is the intersection between sets of pixels. See~\cref{fig:fragmentation} for an example.

Let $B_i = \{b_{i,1}, \dots, b_{i,n}\}$ be the collection of blobs segmented from the $i$th frame of a video $\mathcal{V}$. For every blob $b_{i,j}\in B_i$ we derive the collections of blobs overlapping with $b_{i,j}$ in frames $(i - 1)$ and $(i + 1)$. We call these collections the sets of previous and next blobs of $b_{i,j}$, denoted by $P_{b_{i,j}}$ and $N_{b_{i,j}}$, respectively.

Let $b$ be a blob. We say that $b$ is a blob associated with a \textit{sure individual image} if:

\begin{enumerate}[a)]
\item $b$ is an individual according to~\cref{eq:model_area};
\item $|P_b| = |N_b| = 1$, \ie the blob is overlapping with one and only one blob both in the previous and subsequent frame. The notation $|\cdot|$ indicates the cardinality of a set,~\ie the number of elements of the set.
\item for every $b_p$ and $b_n$ in the past and future overlapping history of $b$ $|P_{b_p}| \leq 1$ and $|N_{b_n}| \leq 1$.
\end{enumerate}

\noindent Symmetrically, we say that $b$ is associated with a \textit{sure crossing image} if:

\begin{enumerate}[a)]
\item $b$ is a crossing according to~\cref{eq:model_area};
\item $|P_b| > 1$ or $|N_b| > 1$.
\end{enumerate}
or
\begin{enumerate}[a)]
\item $b$ does not satisfy the model of the area;
\item $|P_b| = |N_b| = 1$;
\item for some $b_p$ and some $b_n$ in the past and future overlapping history of $b$ $|P_{b_p}| > 1$ and $|N_{b_n}| > 1$.
\end{enumerate}

\subsubsection{Deep crossing detector\label{sec:crossing_detector}}

The methods described in~\cref{sec:model_area,sec:segmented_images_overlapping} can be used on any video in order to generate a dataset $\mathcal{D}_{ic}$ of \textit{sure individual} and \textit{sure crossing} images. With this dataset we train a CNN in the task of distinguishing crossing and individual images. We call this model deep crossing detector (DCD). In the following paragraphs we will describe the preprocessing, architecture, hyperparameters and stopping criteria used to define and train this particular model.

\paragraph{Preprocessing.\label{sec:dcd_preprocessing}} Let $b$ be a blob segmented from a video $\mathcal{V}$ and $I_b$ the image generated by cropping a rectangular bounding box around the centroid of $b$, such that all the pixels of $b$ are represented in $I_b$. We first consider a dilation $b^\star$ of $b$ generated with a $5\times 5$ kernel. We assign value $0$ to every pixel in $I_b$ which is not in $b^\star$. In order to overcome the sensitivity of CNNs with respect to rotation, we compute the first principal component of the cloud of pixels defined by $b$ and then rotate and crop $I_b$ such that the first principal component forms and angle of $\frac{\pi}{4}$ with the x axis.
After the rotation, the size of each image is set to be the maximum of the largest side for all the bounding boxes among the collection of sure crossing images. Then, the images are resized to $40\times 40$ pixels. The resizing improves both the time and memory efficiency of the algorithm. Finally, each image $I\in\mathcal{D}_{ic}$ is standardised as
$I_s = \frac{I - \mu(I)}{\sigma(I)}$ (see sample images in Fig 1, Panel d in the main text).

\paragraph{Architecture.\label{sec:dcd_architecture}} See~\cref{tab:main_cnn_architectures} (deep crossing detector). Both convolutional and fully-connected layers are initialised using Xavier initialisation~\cite{glorot2010understanding}. Biases are initialised to $0$.

\paragraph{Loss function.\label{sec:dcd_loss}} Let $\left(x,l_i\right)$ be a labelled image, where $l_i$ is the label in one-hot encoding,~\ie $l_0 = [1,0]$ is the label associated with $x$ if $x$ is a crossing image, and $l_1 = [0,1]$ if $x$ is an individual.  We compute the loss function associated to $\left(x,l_i\right)$ as a weighted cross-entropy:
\begin{equation}\label{eq:weighted_cross_entropy}
\mathcal{L}\left(x, l_i\right) =-w_i \sum_{j=0}^{n-1} l_i(j) \log s\left( a_j \right)= -w_i l_i(i) \log s\left( a_i \right),
\end{equation}
where $s\left( a_i \right) = \frac{e^{a_i}}{\sum_{j=0}^{n-1} e^{a_j}}$ is the softmax function applied to the activation $a_i$ of the $i$th unit of the last layer of the network, with $j$ varying among all classes, in this case $j\in\Set{0,1}$; $w_i$ is the weight associated with $l_i$ and computed as
\[
w_i = 1 - \frac{\left| L_i \right|}{\sum_{j = 0}^{n-1} \left|L_j\right|},
\]
where $\left| L_i \right|$ is the number of training samples belonging to the class $l_i$ and $j$ varies in the set of all the classes of the dataset (only two in this case: individual and crossing).
The weighting allows to deal with the potentially unbalanced dataset $\mathcal{D}_{ic}$. Indeed, we prefer to collect all the possible examples of sure-crossing and sure-individual images available in a given video, rather than force $\mathcal{D}_{ic}$ to be balanced in the number of samples per class. After dividing the dataset $\mathcal{D}_{ic}$ in batches of $X_i$ of $50$ images, we optimise by considering the mean $\mu\left(\mathcal{L}\left( X_i \right)\right)$ using the algorithm described~\cite{kingma2014adam}, with the hyperparameters suggested in the paper. The learning rate is set at the initial value of $0.005$.

\begin{rem}(On the softmax function)
  In general, the softmax is equipped with an extra parameter called temperature. We omit discussing it in the formula, since we always set it to $1$.
\end{rem}

\paragraph{Training and validation set.\label{sec:dcd_train_val}} Before training, the dataset of \textit{sure crossing} and \textit{sure individual} images is split into two parts: $90\%$ of the images are used for training,~\ie the weights of the network are updated in order to minimise the error (loss function) with respect to the labels associated with this set of images. We call this portion of the dataset the training set, denoted by $T$. The remaining $10\%$ of images--the validation set $V$--are used to evaluate the generalisation power of the network. For this reason, the performances of the model on the validation set are used to stop its training. In~\cref{sec:dcd_stop}, we shall discuss in more detail the algorithm used to stop the training of the network.

\paragraph{Accuracy.\label{sec:dcd_acc}} We measure the accuracy of the network by comparing the predictions generated by the softmax computed on the activation of the last layer of the network, with the labels associated with each image in both the training and the validation set. Hence, let $\left|V\right|$ be the number of images in the validation set, $P_V = \left\{ p_1, \dots, p_n\right\}$ the ordered predictions generated by a forward pass of these images in the network, and $L_V= \left\{ l_1, \dots, l_n\right\}$ the corresponding labels. Let $A_V$ be the set of \textit{correct predictions}, defined as $A_V = \left\{p_i  \mbox{ s.t. } p_i = l_i \mbox{ for }p_i\in P_V,\, l_i\in L_V \right\}$. We define the overall accuracy of the network as
\begin{equation}\label{eq:accuracy}
Acc_V = \frac{\left| A_V \right|}{\left|V\right|}.
\end{equation}
We will also take into account the accuracy on each of the inferred classes. Let $c^\star$ be a class (in this case $c^\star$ could be either the crossing or the individual class). The set
\[
A_V\left(c^\star\right) = \left\{p_i \mbox{ s.t. } p_i = l_i = c^\star \mbox{ for }p_i\in P_V,\, l_i\in L_V\right\},
\]
corresponds to the predictions equal to their associated labels and attributed to the particular class $c^\star$. In this case the class-accuracy is defined as
\begin{equation}\label{eq:class_accuracy}
Acc_V\left(c^\star\right) = \frac{\left| A_V\left(c^\star\right) \right|}{\left|\left\{l\in L_V \mbox{ s.t. } l = c^\star\right\}\right|}.
\end{equation}

\noindent Symmetrically, we define the error and the class-error as $1- Acc_V$ and $1 - Acc_V\left(c^\star\right)$, respectively.

\paragraph{Training stopping criteria.\label{sec:dcd_stop}} While training the network, we verify the goodness of its outputs by computing both the loss function and the accuracy on the validation set $V$ (see~\cref{sec:dcd_loss,sec:dcd_acc}). This procedure gives a reasonable control on the actual classification power of the network on new unlabelled images. Thus, it is crucial to stop the training of the network to prevent two main behaviours. On the one hand, we want to prevent overfitting: A too exact representation of the trainig data, that will prevent the network from generalising on new data points. On the other hand, it is desirable to stop the training in case the error cannot be further minimised, \ie the loss function reached a plateau.

More formally, we call an epoch a complete training pass on the set $T$, concluded with the evaluation (of both loss and accuracy functions) on the validation set $V$. Let $\mathcal{L}_i\left(T\right)$ and $\mathcal{L}_i\left(V\right)$ be the value of the loss function after the epoch $i$. We define
\begin{equation}\label{eq:loss_difference}
d(i^\star) = \mathcal{L}_{i^\star}\left( V \right) - \mu\left(\left\{\mathcal{L}_j\left( V \right) \right\}_{{i^\star}-10 \leq j <{i^\star}}\right)
\end{equation}
as the difference between the loss value in validation at $i^\star$, and the mean of the loss values of the previous $10$ epochs.
We stop the training at epoch $i^\star > 10$ if one of the following conditions holds.
\begin{enumerate}[a)]
                \item The network is overfitting: $d_{i} > 0$ for every $i^\star - 5 < i \leq i^\star$;
                \item the loss reached a plateau: $\left|d_{i^\star}\right| < 0.05 \cdot 10^{\log_{10}\left( \mathcal{L}_{i^\star}\left( V \right)\right) - 1}$;
                \item the network reached class-accuracy $1$ on all the classes, for every sample in the validation set. See~\cref{eq:accuracy};
                \item the loss (error) is zero: $\mathcal{L}_{i^\star}\left( V \right) = 0$.
\end{enumerate}

\paragraph{Crossing detection.} Let $\Delta$ be the set of parameters learnt by training the DCD as described above. Let us denote the trained model as $\mbox{DCD}\left(\Delta\right)$. We create the test set $\mathbb{T}$ of unlabelled images by considering all the images that are not either \textit{sure individuals} or \textit{sure crossings}. The trained model acts as a function taking as an input an image $I\in\mathbb{T}$ and outputting a prediction as the softmax computed on the activation of the last layer of $\mbox{DCD}\left(\Delta\right)$. We recall that the softmax is the function defined as
\begin{equation}\label{eq:softmax}
  s\left(a_i\right) = \frac{e^{a_i}}{\sum_j e^{a_j}},
\end{equation}
where $a_i$ is the activation of the $i$th unit of the last layer. Since the DCD classifies images in two classes, its last layer is composed by two units. Hence, given an image $I$, we obtain
\[
\mbox{DCD}\left(\Delta\right)\left(I\right) = \Set{s_1, s_2},
\]
where for brevity we set $s_i = s\left(a_i\right)$. If $s_1 > s_2$ the image is classified as a crossing, and as an individual otherwise.

\paragraph{Exceptions}
It is possible that during training the loss value diverges to infinity. In this case a warning is produced, and the algorithm falls back into a crossing-individual images discrimination process based only on model of the area of individual blobs (see~\cref{sec:model_area}). In case the criteria forcing the training to stop are never reached, we set a maximum number of $100$ epochs for the training of the DCD. If this threshold is reached, a warning is produced and the training is stopped. The parameters computed in the last iteration will be used to classify individual and crossing images.

\subsection{Fragmentation\label{sec:fragmentation}}

\begin{figure}[tb]
\centering
\includegraphics[width=.6\textwidth]{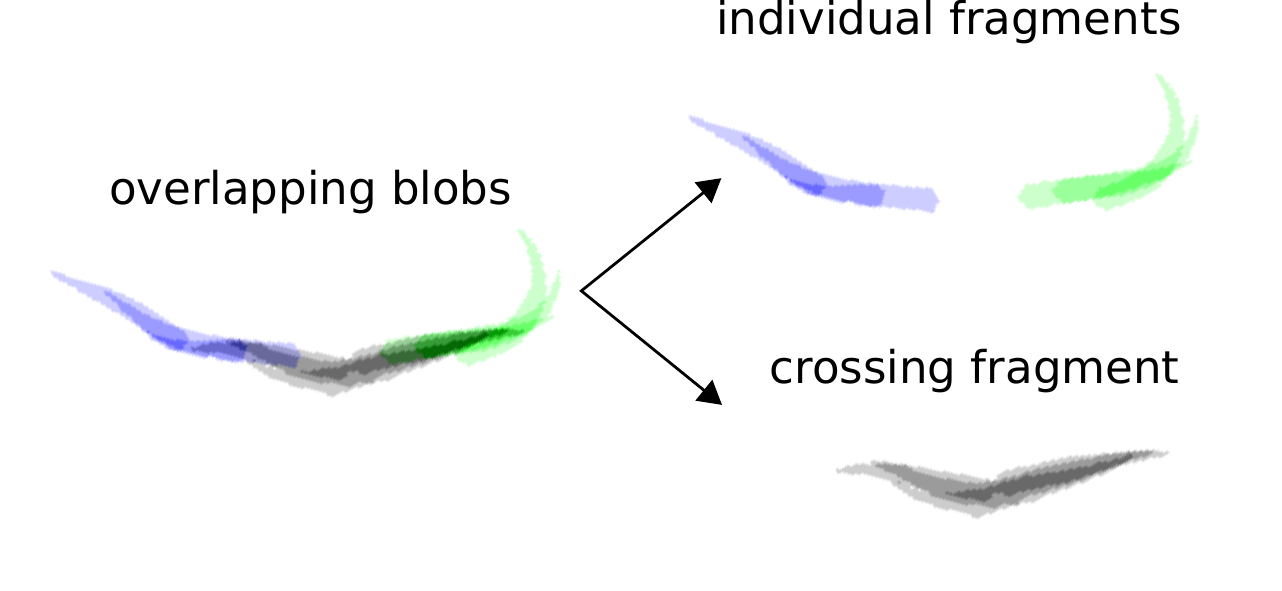}
\caption{An individual fragment built by considering the blobs' overlapping in subsequent frames.} \label{fig:fragmentation}
\end{figure}

At this stage of the algorithm, the images segmented from the video (see~\cref{sec:segmentation}) are labelled either as individual or crossing, following the protocols described in~\cref{sec:crossing_detection_CNN}. A very careful dynamical analysis of the segmented blobs allows to create collections of images associated with the same individual (or crossing) in subsequent frames. In the remainder, we will refer to these collections as \textit{individual} and \textit{crossing} \textit{fragments}. See~\cref{fig:fragmentation} for an example of fragments and its decomposition in individual and crossing components.

The method used to create these types of fragments is based on the one hand on the classification of the images into crossing and individual categories; on the other hand, it considers the overlapping of the blobs associated with these images. We start by introducing some notations. Then, we will describe the algorithm to generate individual and crossing fragments in two separated sections.

Let
$\mathcal{B} = \left\{b_{1,1}, b_{1,2}, \dots b_{1,n_1}, \dots b_{m,1}, b_{m,2}, \dots, b_{m, n_k} \right\}$ be the collection of segmented blobs, where the first index of the elements of $\mathcal{B}$ corresponds to the frame number, and the second to the order in which the blobs have been segmented. We recall, that given two blobs $b_1$ and $b_2$, we say that they overlap if and only if the intersection of their sets of pixels is not empty. Following the notation introduced in~\cref{sec:segmented_images_overlapping}, given a blob $b_{i,j}\in\mathcal{B}$ we call the collections of blobs overlapping with $b_{i,j}$ in frame $(i-1)$ and $(i+1)$, $P_{b_{i,j}}$ and $N_{b_{i,j}}$ respectively.

\subsubsection{Individual fragments\label{sec:individual_fragments}}
We iterate over the elements of $\mathcal{B} = \left\{b_{i,j}\right\}$ proceeding by frame number $i$ and then following the natural ordering induced by the second index. Let $b_{i,j}$ be a blob associated with the image $I_{b_{i,j}}$ labelled as an individual. We create individual fragments by considering only the future overlapping history of $b_{i,j}$. If $b_{i,j}$ is not yet part of any individual fragment, we associate with $b_{i,j}$ a unique fragment identifier $\alpha$ (\ie $b_{i,j}$ is the blob intiating an individual fragment). To simplify the notation let $b_{i,j} = b$, and $N_{b_{i,j}} = N$. We consider two cases:
\begin{enumerate}[ {case} 1:]
	\item $|N|>1$. The blob $b$ in frame $i$ overlaps with more than one blob in frame $i+1$, hence it is the only blob (and image) associated with the individual fragment $\alpha$.
	\item $|N| = 1$. Let $n_b$ be the unique element of $N$. The fact that $b$ overlaps with a single blob in its future history is a necessary condition for $n_b$ to be part of the same fragment as $b$, but not sufficient. It could be that $|P_{n_b}|>1$, thus we say that $n_b$ is in the same individual fragment as $b$ if and only if $|P_{n_b}| = 1$. We also require the image $I_{n_b}$ to be labelled as an individual.
\end{enumerate}
If case 1 is verified we generate a new fragment identifier and continue iterating on the elements of $\mathcal{B}$. Otherwise, we apply the same algorithm to $n_b$ in order to enlarge the individual fragment $\alpha$ as much as possible. We stop adding blobs to the fragment whenever, during the iteration, a new candidate blob $n_{b_{last}}$ fulfils the condition in case 1. See~\cref{alg:individual_fragment_identiftier} for the pseudocode.

\begin{algorithm}[tb]
\caption{Assign individual fragment identifier to blobs}\label{alg:individual_fragment_identiftier}
\begin{algorithmic}[1]
\Procedure{Assign\_fragment\_identifier}{$\mathcal{B}$}
\State $\alpha = 0$
\For{$b\in\mathcal{B}$}
\State $cur_b = b$
\If{$cur_b$ has no fragment identifier and $I_{cur_b}$ is an individual image}
\State compute $N_{cur_b}$
\If {$|N_{cur_b}| = 1$}
\State compute $P_{n_b}$ for $n_b\in N_b$
\If{$|P_{n_{cur_b}}| = 1$ and $I_{n_{cur_b}}$ is an individual image}
\State $n_{cur_b}$ is part of the fragment $\alpha$
\State $cur_b = n_b$
\While{$|N_{cur_b}| = 1$, $|P_{n_{cur_b}}| = 1$ and $I_{n_{cur_b}}$ is an individual image}
\State $n_b$ is assigned to the fragment $\alpha$
\State $cur_b = n_b$
\EndWhile
\Else{}
\State $\alpha = \alpha + 1$
\EndIf
\Else
\State $\alpha = \alpha + 1$
\EndIf
\Else{}
\State continue
\EndIf
\EndFor
\EndProcedure
\end{algorithmic}
\end{algorithm}

\subsubsection{Crossing fragments\label{sec:crossing_fragments}}
In the same setting of the previous section let $b_{i,j} = b$ be a blob associated with a crossing image. If $b$ is not yet equipped with a crossing fragment identifier, we generate a new identifier $\beta$. The conditions are almost equivalent to the individual fragments' case:
\begin{enumerate}[{case} 1:]
	\item $|N|>1$. The blob $b$ in frame $i$ overlaps with more than one blob in frame $i+1$, hence the crossing represented by $b$ is splitting. Thus $b$ is the only blob associated with the crossing fragment $\beta$.
	\item $|N| = 1$ and $|P_{n_b}| = 1$ and $I_{n_b}$ is a crossing image. We add $n_b$ to the crossing fragment $\beta$.
\end{enumerate}
In the second case, we try to extend the crossing fragment simply by iterating the algorithm on $n_b$, and verifying that both $P\left(n_b\right)$ and $N\left(n_b\right)$ have cardinality $1$, and the unique element of $N\left(n_b\right)$ is associated with a crossing image.

The pseudocode presented in~\cref{alg:individual_fragment_identiftier} can be easily adapted to work with crossing fragments, by modifying the \textit{if}s and while conditions.

\subsection{Cascade of training/identification protocols\label{sec:deep_protocol_cascade}}
After fragmentation has finished, the training of the identification network begins. We would first like to give the reader some intuition regarding why it is possible to train an identification network in an automated manner. First imagine that we had at our disposal an all-knowing black box, that looked at the set of fragments we have complied from one part of the video and then told us which fragment belonged to which individual. Remember that each fragment contains an entire set of images belonging to a single individual. Therefore, thanks to the information coming from the black box, we would effectively have at our disposal a set of labelled images and we could use standard supervised learning to train a classifier which can tell the individuals apart from one another. The trained network could then be used on other parts of the video to do identification.

In real life, we do not have access to such source of information, so we need to use some heuristics to generate our training dataset. In order to understand our heuristic, let's again remember that each fragment is supposed to contain images belonging to a single individual. We also know the total number of individuals in the video as this number is specified by the user. Consequently, if we can find one frame of the video, where the number of fragments which are present at that frame is equal to the total number of individuals, then we could be sure that each visible fragment in that frame belongs to a separate individual. We can then label the images within each fragment with the same label while every fragment of course has a different label. Next, we train our network on the resulting dataset of images and labels. This is the intuition behind how we achieve our aim without the help of an omniscient black box.

In order to train the identification network, we have designed three training protocols. The first protocol is the fastest and is able to deal with videos where animals are relatively well separated (crossings are not too frequent). The other two protocols handle more difficult scenarios, where crossings may be frequent, the setup lighting intensity may drift over time, the animals may change their features throughout the video (\eg,~posture, colour).

Each protocol relies on the information acquired and structures defined in the previous ones. In the following sections we will introduce some definitions and the main elements on which the fingerprint protocols are built. Then, we will discuss each of the three protocols from the simplest and fastest, to the most general and computationally expensive one.

\subsubsection{Global fragments\label{sec:global_fragments}}
All the protocols rely on a strong, fundamental hypothesis: To learn the features characterizing each individual and consequently identify it, there must exist at least one portion of the video in which all animals are visible and separated.

\begin{figure}[tb]
\centering
\includegraphics[width=.7\textwidth]{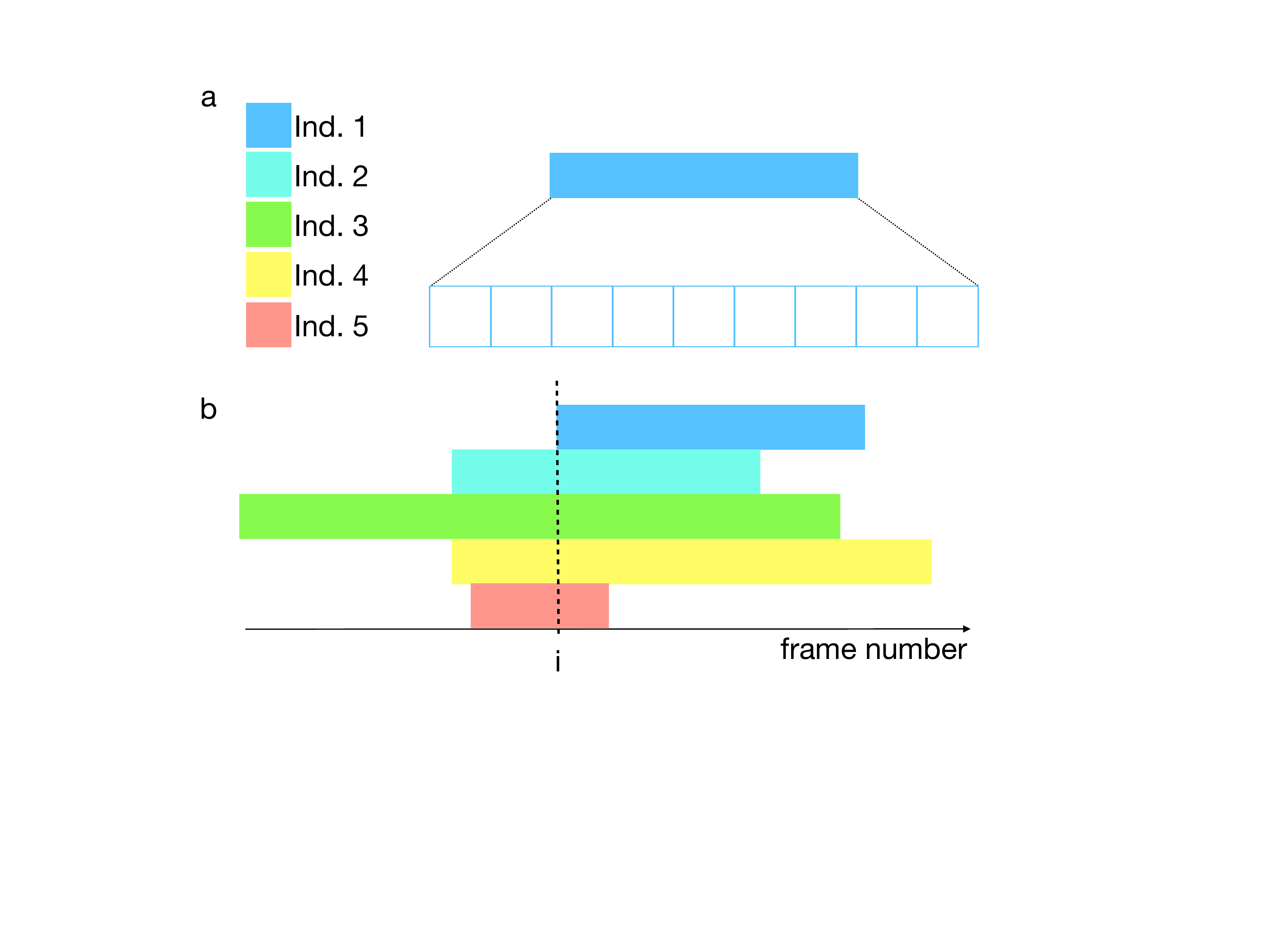}
\caption{Global fragment in a video of $5$ individuals. A) Each individual in the video is associated with a certain label. An individual fragment is a collection of preprocessed images associated with the same individual. B) A global fragment is a collection of individual fragments that coexist at least in a frame of the video, the frame $i$ in the image.\label{fig:global_fragments}}
\end{figure}

Let $\mathcal{V}$ be a video in which the aforementioned condition is fulfilled in frame number $i$. We define a \textit{global fragment} as the collection of individual fragments (see~\cref{sec:individual_fragments}), whose images contain the ones extracted from the $i$th frame of the video and with the same number of fragments as the number of individual to be tracked. We call the minimum frame number in which this condition is satisfied the \textit{core of the global fragment}. See~\cref{fig:global_fragments} for a visual representation of a global fragment. We denote by $\mathcal{G}$ the set of all the global fragments in $\mathcal{V}$, whose shortest individual fragment counts at least $3$ images.

\subsubsection{Identification network\label{sec:identification_network}}

All the fingerprint protocols aim at finding different strategies in order to create datasets of images of the animals labelled with their identities. These datasets will be created from one, or a collection of global fragments and used to train the \textit{identification CNN}, denoted in the remainder as idCNN. In the following paragraphs we define the architecture, the hyperparameters and algorithms used to train the idCNN.

\paragraph{Preprocessing.\label{sec:idcnn_preprocessing}} The images used to train and test the idCNN are preprocessed with an algorithm similar to the one described in~\cref{sec:crossing_detector}. The images are obtained, aligned and standardised in the same way. The only difference is that the square images used to train the DCD are resized to be square images of size $40\times 40$, while the training images of the idCNN are obtained as $\frac{\mbox{estimated body length}}{\sqrt{2}}$. The body length is estimated by considering the median of the diagonal of the images associated to each individual blob.

\paragraph{Architecture.\label{sec:idcnn_architecture}} See~\cref{tab:main_cnn_architectures} (identification convolutional neural network). Both convolutional and fully-connected layers are initialised using Xavier initialisation~\cite{glorot2010understanding}. Biases are initialised to $0$.

\paragraph{Loss function.\label{sec:idcnn_loss}} The loss is a weighted cross-entropy~\cref{eq:weighted_cross_entropy}.
The dataset given by a global fragment is potentially unbalanced: Every individual fragment $F_i\in G$ counts a certain number of images, say $n_i$. For every $F_i\in G$, we compute the weight $w_i$ of~\cref{eq:weighted_cross_entropy} as
\[
w_i = 1 - \frac{n_i}{\sum_j n_j},
\]
 where $j$ varies in $\Set{F_1,\dots, F_n} \in G$. Thus, a larger loss is associated with individuals less represented in the dataset. We optimise using stochastic gradient descent, setting the learning rate to $0.005$.

\paragraph{Training and validation set.\label{sec:idcnn_train_val}} Every individual fragment $F_i\in G$ can be written as $F_i = \left\{\left(I_1, l_1 \right), \dots,  \left(I_{n_i}, l_{n_i} \right)\right\}$, where $\left(I_j, l_j \right)$ is a pair such that the label $l_j$ is the identity of the individual depicted in the image $I_j$.
The dataset generated from $G$ is given by the union $\mathcal{D}_G = \cup_i F_i$. After a random permutation of the pairs $\left(I_j, l_j \right)$ in order to lose any temporal correlation between the images, we split $\mathcal{D}_G$ in the training and validation sets denoted by $T$ and $V$ respectively. These sets are composed of $90\%$ and $10\%$ of the available data, respectively.

\paragraph{Accuracy.\label{sec:idcnn_acc}} The accuracy of the network is computed as the number of successfully classified images over the total number of images, according to~\cref{eq:accuracy}. We measure the single class accuracy following~\cref{eq:class_accuracy}. This second expression is fundamental when dealing with large groups, in order to evaluate the capability of the network to distinguish each of the individuals.

\paragraph{Training stopping criteria.\label{sec:idcnn_stop}} See~\cref{sec:dcd_stop}.

\paragraph{Exceptions}
It is possible that during training, the loss value diverges to infinity. In this case an error is raised and the algorithms stop its execution. Advanced users have the possibility to change the parameters of the idCNN (\eg learning rate, dropout, layers' units count).

If the training of the idCNN is not stopped before $10000$ epochs (passes over the entire dataset) a warning is produced and the training is stopped. The parameters computed in the last epoch will be used to continue the fingerprint protocol cascade.

\begin{figure}[tb]
\centering
\includegraphics[width=.5\textwidth]{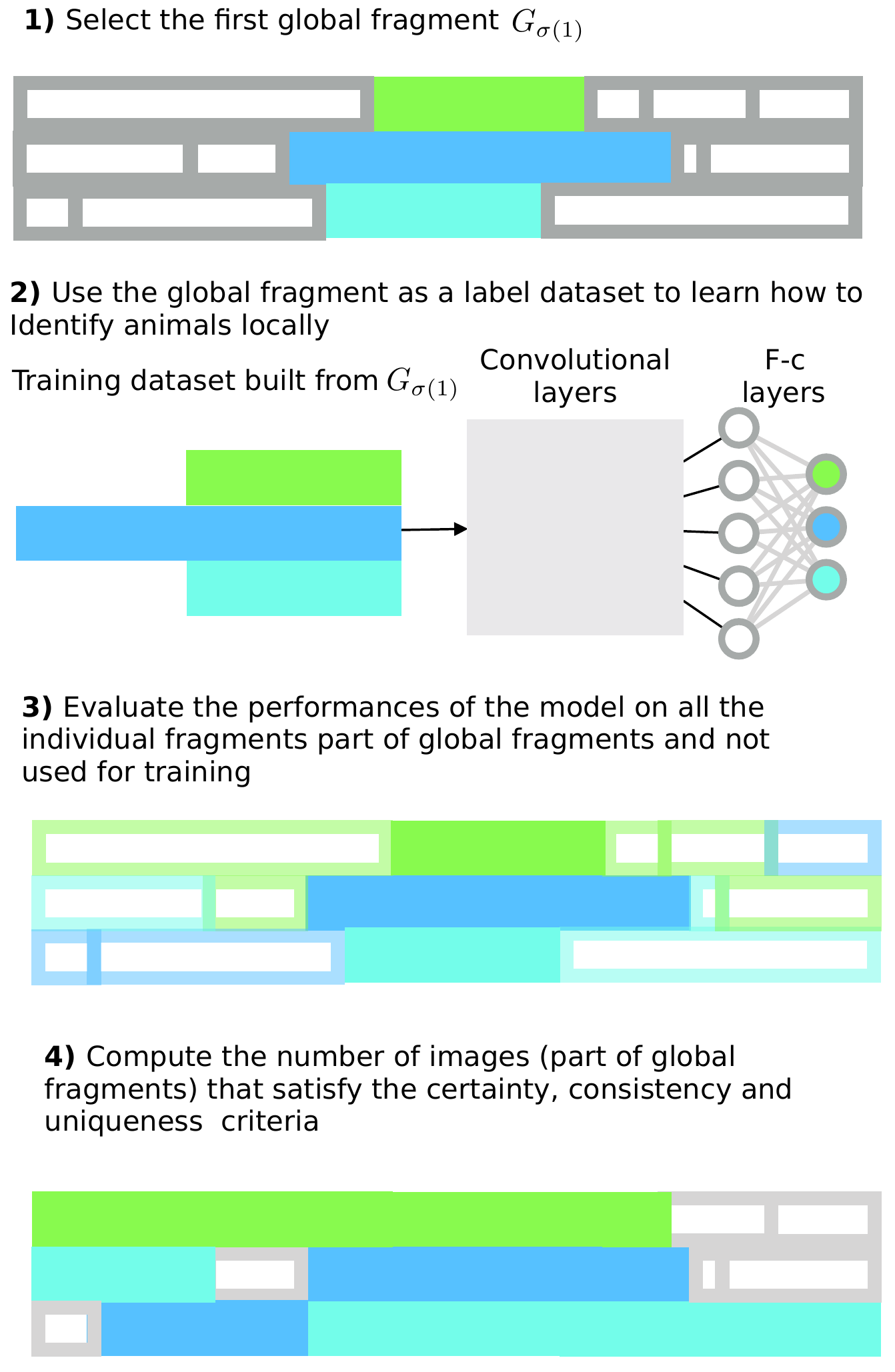}
\caption{\textbf{Protocol 1.} The simplest fingerprinting protocol takes advantage of the information of a single global fragment (1) chosen according to the algorithm discussed in~\cref{sec:choose_first_global_fragment}, in order to train the idCNN (2) (see~\cref{sec:idcnn_train_val}). In step (3), the entire collection of individual fragments belonging to at least one global fragment is identified by using the classification provided by the model trained in (2) and according to the individual fragments identification algorithm described in~\cref{sec:individual_fragments_identification_1}. Finally, the performace of the model is evaluated (see~\cref{sec:idcnn_model_evaluation}).} \label{fig:protocol1}
\end{figure}

\subsubsection{Protocol 1: One-global-fragment tracking\label{sec:protocol_one}}
 This protocol is based on the features learnt by considering the images belonging to a single global fragment. Thus, it is likely to be successful when the individual images are uniform along the entire video.

\paragraph{Choosing the global fragment.\label{sec:choose_first_global_fragment}} Since the network will be trained on a single global fragment, its choice is fundamental. We aim at selecting the global fragment whose individual fragments are sets of images with high variability. This, in order to be as close as possible to the setting described in~\cref{fig:Single_image_identification_accuracy_for_different_CNN_architectures}, where images are subsampled from the entire video, and hence uncorrelated in time.

We define the distance travelled in a global fragment $G$ as the minimum of the distance travelled in its individual fragments, as described in~\cref{alg:get_distance_travelled}.

We choose the global fragment realising the maximum of the minimum distance travelled denoted as $G_{\sigma(1)}$. This procedure does not assure to get the global fragment whose images are maximally variable. However, there is a natural correlation between the distance travelled by an animal and the variability of the images stored in the corresponding fragment.

\begin{algorithm}[tb]
\caption{Compute distance travelled in individual fragment}\label{alg:get_distance_travelled}
\begin{algorithmic}[1]
\Procedure{compute\_distance\_travelled}{$F$}
\State $c\left(F\right) = \{c_1, \dots, c_n\}$\Comment{Centroids of the images}
\State distance\_travelled $= 0$
\For{$\left(c_i, c_{i+1}\right) \in F\setminus\{c_n\}\times F$}\Comment{For each pair of subsequent blobs}
\State $d = \parallel c_{i+1}, c_{i}\parallel_2$\Comment{Compute the Euclidean distance}
\State distance\_travelled $+= d$\Comment{Add it to the overall distance}
\EndFor
\State \textbf{return} distance\_travelled
\EndProcedure
\end{algorithmic}
\end{algorithm}

\paragraph{Training.} After choosing $G_{\sigma(1)}$, we label the images belonging to each of its individual fragment with random, unique identities (think of increasing natural numbers). From this dataset of labelled images we create the training and the validation set, as described in~\cref{sec:idcnn_train_val}. We train the idCNN as specified in~\cref{sec:idcnn_loss}. The training is interrupted automatically when one of the condition in~\cref{sec:idcnn_stop} is satisfied. Let us call $\theta_0$ the set of parameters of the idCNN after training, and denote the trained model as idCNN$\left(\theta_0\right)\left(I\right)$.

\paragraph{Single image identification.\label{sec:single_image_identification}}
We recall that a trained neural network acts as a function. We use the trained idCNN in order to identify individual fragment not used for training. For every image $I$ in an individual fragment $F$, we can compute idCNN$\left(\theta_0\right)\left(I\right)$, obtaining $S_I = \Set{s_1, \dots, s_n}$, where
\[
s_j = \frac{e^{a_j}}{\sum_i e^{a_i} }
\]
is the softmax computed on the activity of the $j$th unit of the last fully connected layer. The idCNN's last layer has as many units as the number of individual to be tracked (see~\cref{tab:main_cnn_architectures}, idCNN). By definition, $\sum_{s\in S_I} s = 1$. Thus, $S_I$ can be interpreted as the probability of the input image to represent each of the individuals. Each image is labelled with the identity realising the maximum of the softmax: $id\left(I\right) = \argmax\left(S_I\right)$.

\begin{rem}
Given the relatively low number of parameters of  idCNN ($\approx 200000$, see~\cref{tab:main_cnn_architectures}, idCNN), and the usage of GPU computing, the single image identification step is time efficient, even when dealing with large groups of animals.
\end{rem}

\paragraph{Computing the identity probability mass function.\label{sec:individual_fragments_identification_1}} When considering an individual fragment, it is natural to take advantage of the hypothesis that all its images are associated with the same individual. We follow the assumptions and logic already presented in \cite[Supporting text, Section 3.1]{Perez-Escudero2014}. Let $\Lambda_F = \left(f_1, \dots, f_n\right)$ be the vector of identification frequencies associated with $F$, \ie the vector whose components correspond to the number of images of $F$ assigned to the $i$th individual, computed as in~\cref{alg:compute_freqs}. Under the assumption that all the images in $F$ are independent and that the probability to assign one image to the correct individual is twice
as large as the probability to assign the image to any of the incorrect individuals, we compute for every identity $i\in\Set{1, \dots, n}$

\begin{equation}\label{eq:P1}
P_1\left(F, i \right) = \frac{2^{\Lambda_F(i)}}{\sum_{i = j}^n 2^{\Lambda_F(j)}},
\end{equation}
where $\Lambda_F(i)$ is the $i$th component of the vector $\Lambda_F$.

\noindent The vector $P_1\left(F\right) = \left(P_1\left(F, 1\right), \dots, P_1\left(F, n\right)\right)$ is the probability mass function of $F$ being identified as one of the individuals.

\paragraph{Model quality check and identification of individual fragments in global fragments.\label{sec:idcnn_model_evaluation}}
While identifying the individual fragments belonging to the global fragments not used to train the idCNN$\left(\theta_0\right)$, we evaluate the goodness of the model. For every $G\in\mathcal{G} \setminus \Set{G_{\sigma(1)}}$ we proceed by verifying the following conditions, providing at the same time a temporary identification of the individual fragments in $G$:

\begin{enumerate}[1.]
  \item \textbf{$G$ is certain:} A global fragment $G$ is certain if all the individual fragments $F_i$ in $G$ are also certain. A fragment $F$ is certain if $cert(F) \geq 0.1$. Let $a$ and $b$ be the indices (identities) realising the first and second maximum of $P_1\left(F\right)$ respectively, and $S_j$ the vector of softmax values of all the images assigned to the index $j$ in the fragment $F$. The function $cert\left(F\right)$ is defined as

  \begin{equation}\label{eq:certainty}
  cert\left(F\right) = \frac{\mbox{median}\left(S_{a}\right) \cdot P_1\left(F, a\right) - \mbox{median}\left(S_{b}\right) \cdot P_1\left(F, b\right)}{ P_1\left(F, a\right) + P_1\left(F, b\right)}.
  \end{equation}

  \item \textbf{Temporary identification of the individual fragments:} Let us consider the entire collection of individual fragments $\left\{F_i\right\}_{F_i\in G}$. We start by reordering it according to the maximum value of each $P_1\left(F_i\right)$. Let us denote the reordered collection of individual fragments as $\mathcal{F} = \Set{F_{\rho(1)}, \dots, F_{\rho(m)}}$. We iterate on the individual fragments indexed as in $\mathcal{F}$. So, $F_{\rho(1)}$ is the individual fragment having maximum probability of being identified as the individual with identity $\iota = \argmax\left(P_1\left(F_{\rho(1)}\right)\right)$. We set $\iota$ to be the temporary identity of $F_{\rho(1)}$ if two conditions are reached:
  \begin{enumerate}[1.]
    \item There is no identified individual fragment coexisting with $F_{\rho(1)}$ with identity $\iota$.
    \item $\max_{f_j\in P_1\left(F_{\rho(1)}\right)}\left(P_1\left(F_{\rho(1)}\right)\right) > \frac{1}{\left|F_{\rho(1)}\right|},$ where $\left|F_{\rho(1)}\right|$ is the number of images in $F_{\rho(1)}$.
  \end{enumerate}
  We proceed to the next iteration by considering $F_{\rho(2)}$. If one of the aforementioned conditions is not satisfied, the individual fragment is marked as \textit{non-consistent} as well as the entire global fragment $G$.

  \item \textbf{$G$ is unique:} A global fragment $G$ is unique if the temporary identity of every $F_i$ in $G$ are unique within the global fragment.
 \end{enumerate}

\noindent We iterate on the global fragments in $\mathcal{G} \setminus \Set{G_{\sigma(1)}}$ by sorting them from the nearest to the farthest with respect to the distance in frames between their core and the core of $G_{\sigma(1)}$. See~\cref{sec:global_fragments} for the definition of the core of a global fragment.

We now consider the number of images in all the individual fragments that are part of a global fragment. We will refer to this set of images in the remainder as the \textit{global fragments' images} or the \textit{images in global fragments}. If at least $99.95\%$ of these images are contained in global fragments considered acceptable with respect to the conditions listed in the previous paragraph, we interrupt the cascade of protocols and we pass to the residual identification described in~\cref{sec:final_identification}. Otherwise, the second protocol is put in place.

\begin{algorithm}[tb]
\caption{Compute identification frequencies in individual fragment}\label{alg:compute_freqs}
\begin{algorithmic}[1]
\Procedure{get\_frequencies}{$F$}\Comment{An individual fragment}
\State $S_F = []$ \Comment{softmax array}
\State $ids = []$ \Comment{identities array}
\For{ $I\in F$}\Comment{For every image in $F$}
\State $s_I = \softmax\left(\mbox{idCNN}\left(\theta_0\right)(I)\right)$\Comment{Compute softmax (GPU)}
\State $S_F.append(s_I)$
\State $ids.append(\argmax\left(s_i\right))$\Comment{Compute the identity (GPU)}
\EndFor

\Return frequencies = count(ids) \Comment{Count assignment frequency}
\EndProcedure
\end{algorithmic}
\end{algorithm}

\begin{figure}[tb]
\centering
\includegraphics[width=.5\textwidth]{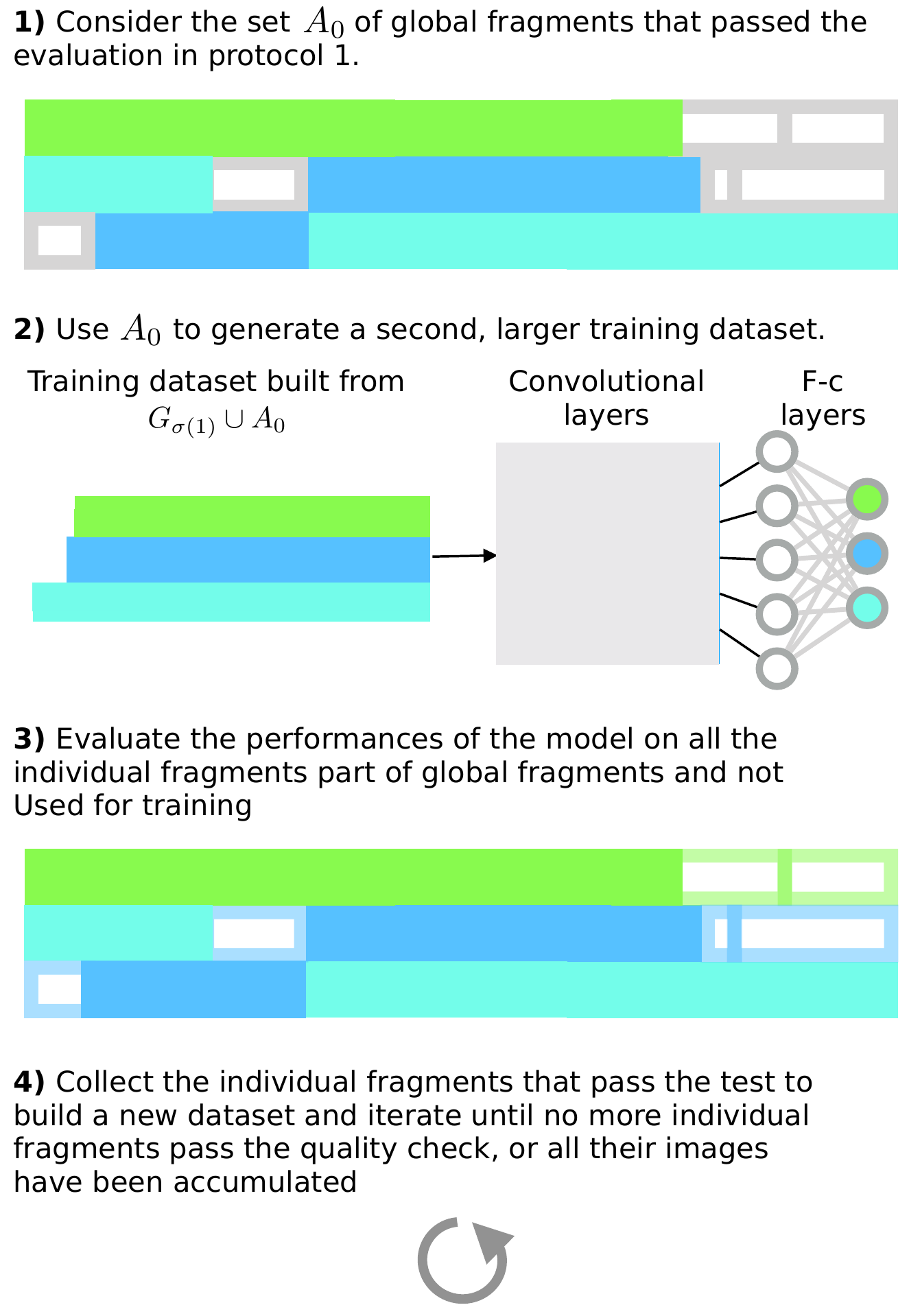}
\caption{\textbf{Protocol 2.} Flow of the $i$th iterative step of protocol 2. In this protocol fingerprints are built iteratively, by iterative accumulation. In (1) images belonging to individual fragments assigned with high certainty in the $i-1$th iteration are collected to form a new, broader dataset of labelled images (see~\cref{sec:accumulation,sec:partial_accumulation}). This dataset is used to fine-tune the idCNN in (2). Individual fragments in the video that still lack identity are identified (3). The most certain ones are used to initialise the next iteration step.} \label{fig:protocol2}
\end{figure}

\subsubsection{Protocol 2: Global-fragments-accumulation\label{sec:protocol_two}}
The main aim of this protocol is to accumulate the images belonging to those global fragments, detected during protocol 1, that are simultaneously certain, consistent and unique. By iterating this procedure, it is possible to incorporate new images in the labelled dataset used to train the idCNN. This accumulation procedure allows to learn features able to grasp the individuals' variability through the video. See~\cref{fig:protocol2} for the flow of the algorithm of this protocol.

\paragraph{Global accumulation.\label{sec:accumulation}} Let $A_{-1} = \Set{G_{\sigma(1)}}$ be the set containing the first global fragment used for training, and $A_0 = \Set{ G_{1,1},\dots, G_{1,n}}\subset \mathcal{G}\setminus G_{\sigma(1)}$
be the subset of global fragments that meet the conditions described in~\cref{sec:idcnn_model_evaluation}.

First, we fix the identities of the individual fragments belonging to the global fragments in $A_0$, since the images associated with these individual fragments will be used to train the idCNN.

We build the dataset $\mathcal{D}_{A_0}$ by considering all the labelled images contained in every global fragments in $A_{-1}\cup A_{0}$.  Note that, since an individual fragment can be shared by more global fragments, its images will be collected only once.

$\mathcal{D}_{A_0}$ is then split in the training and validation sets ($T_1$ and $V_1$), according to the proportions specified in~\cref{sec:idcnn_train_val} and an additional constraint: In the training and validation sets every individual can be represented by at most $3000$ images. If the amount of images associated with a certain individual in $\mathcal{D}_{A_0}$ exceeds this threshold, $3000$ images are randomly subsampled from this collection, by taking $1800$ samples from the images previously accumulated (images in $A_{-1}$), and the remaining $1200$ from the new set of accumulated images ($A_{0}$).

\begin{rem}
  At every iteration of the accumulation, the permutation used to subsample the images representing the same individual changes in order to train the idCNN with maximally variable images.
\end{rem}

We train the network using the stopping criteria listed in~\cref{sec:dcd_stop}. Following the notation introduced in~\cref{sec:protocol_one}, we denote the idCNN model obtained after training as $\mbox{idCNN}\left(\theta_1\right)$. The accumulation process is iterated by defining $A_i$ as the set of acceptable global fragments in $\left\{\mathcal{G}\setminus {G_{\sigma(1)}}\right\}\setminus \cup_{j = 0}^{i - 1}{A_j}$. The set of new acceptable global fragments is computed by identifying the individual fragments not used for training and apllying the procedure described in~\cref{sec:idcnn_model_evaluation}.

\paragraph{Partial accumulation.\label{sec:partial_accumulation}} Partial accumulation is a riskier accumulation strategy. It allows to include in the dataset of accumulated images single individual fragments, rather than entire global fragments. For this reason, before applying this strategy, we require that more than half of the images contained in the set of global fragments has been accumulated via global accumulation. Assume that this condition is reached at the iteration $\bar{i}$, then an individual fragment $F\in G$ for some global fragment $G\not\in A_{\bar{i}}$ is accumulated if
\begin{enumerate}[1.]
	\item $cert\left( F \right) > 0.1$;

  \item let $\gamma\left( F \right)$ be the set of individual fragments that coexist with
  $F$ in at least one frame of the video. $F$ can be accumulated if at least half of the elements of $\gamma\left(F\right)$ have already been accumulated;

	\item the identity of $F$ is coherent with all the individual fragments in $\gamma\left( F \right)$, \ie the assignment of a certain identity to $F$ does not create duplications.
\end{enumerate}
If all these conditions are met, $F$ will be added to $A_{\bar{i} + 1}$ as a single individual fragment.

\paragraph{Accumulation stopping criteria.\label{sec:accumulation_stopping}}
We stop both the global and the partial accumulation processes if one of the two following conditions holds:
\begin{enumerate}
	\item $99.95\%$ of the images in global fragments have been accumulated;
	\item there are no more acceptable global or individual fragments.
\end{enumerate}

\paragraph{Evaluation of the accumulation.\label{sec:accumulation_evaluation}} If the number of images accumulated in the last iteration is less than $90\%$ of all the images in the global fragments, the accumulation is considered not acceptable and the third protocol is used. Otherwise we proceed to the identification of the individual fragments not identified during accumulation, see~\cref{sec:final_identification}.

\begin{figure}[tb]
\centering
\includegraphics[width=.6\textwidth]{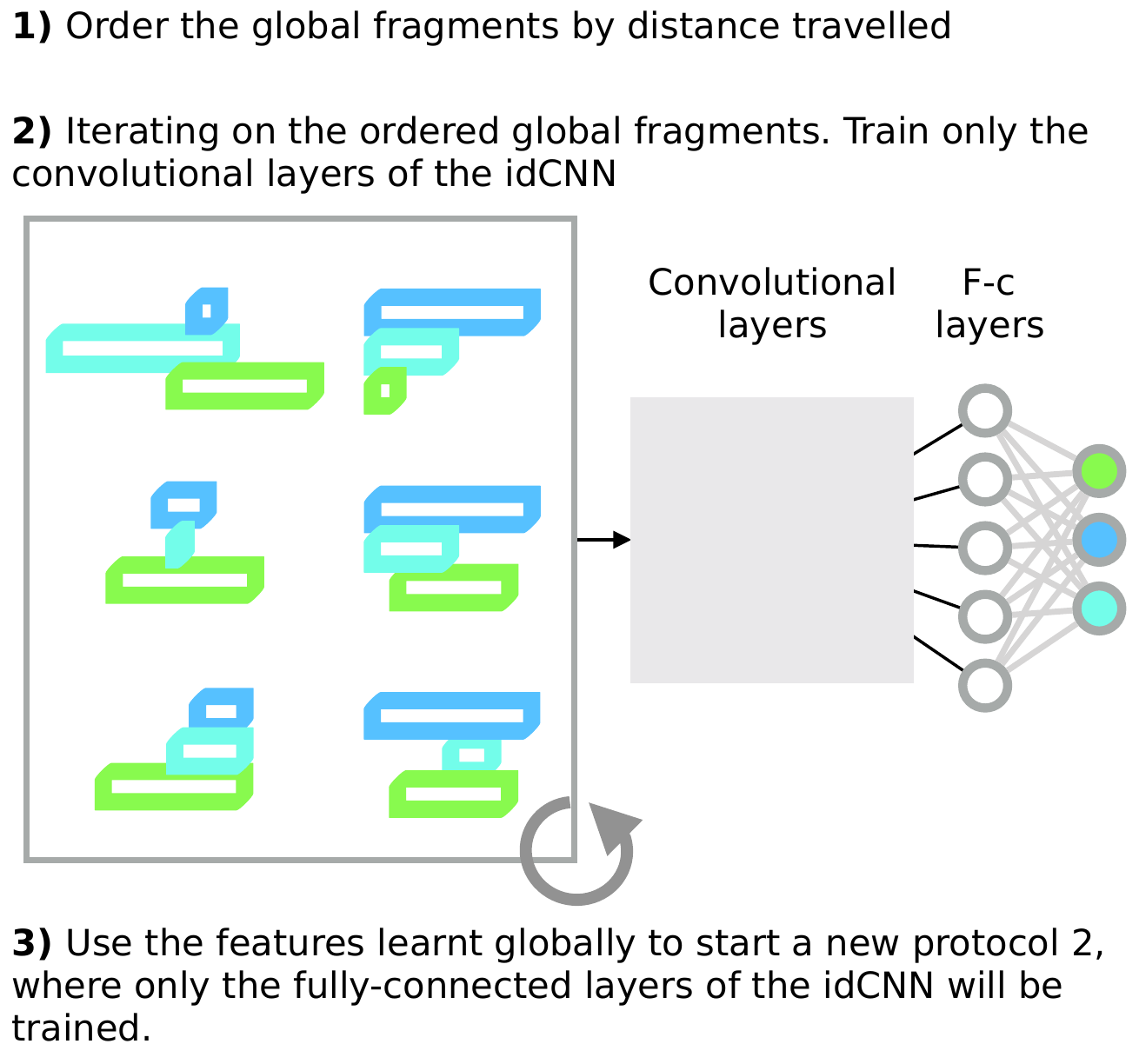}
\caption{\textbf{Protocol 3.} The acquisition of the information that will be used to produce the fingerprints is split in two parts in this protocol. First, we consider the collection of global fragments, by deleting any previous identification (1). In step (2) we train the idCNN iterating on the set of global fragments. When iterating, we keep the weights of the convolutional part, as while the classifier is reinitialised. Finally we re-start protocol 2 by using the encompassing convolutional features learnt form the entire video, and training only the classification part og the network.} \label{fig:protocol3}
\end{figure}

\subsubsection{Protocol 3: pretraining and accumulation\label{sec:protocol3}}
This last protocol allows to learn globally the features of the images representing the individuals by taking advantage of their local organisation in global fragments.

\paragraph{Pretraining.\label{pre_training}} Given a video $\mathcal{V}$ let $\mathcal{G}$ be the set of global fragments as defined in~\cref{sec:global_fragments}. We rewrite the idCNN by considering it as the juxtaposition of its convolutional $\mbox{idCNN}_c$ and fully-connected $\mbox{idCNN}_f$ parts, with sets of parameters $\Gamma$ and $\Phi$, respectively. Here follows the list of processes involved in the pretraining algorithm:
\begin{enumerate}[(i)]
		\item Consider the set $\sigma\left(\mathcal{G}\right) = \Set{G_{\sigma(1)}, \dots, G_{\sigma(n)}}$ of global fragments $\mathcal{G}$ ordered by distance travelled. See~\cref{sec:choose_first_global_fragment}.

    \item Iterate on the elements of $\sigma\left(\mathcal{G}\right)$. Let $\mathcal{D}_{G_{\sigma(i)}}$ be the dataset of labelled images built at the $i$th iteration. Assign a random unique identity to each individual fragment: The aim is to learn features, and classify the individuals only locally. Generate both the training and validation sets as in~\cref{sec:idcnn_train_val}.

    \item Train the model using the parameters learnt during the previous iteration for the convolutional part $\mbox{idCNN}_c\left(\Gamma_{\sigma(i-1)}\right)$ and reinitialise $\mbox{idCNN}_f$. This step allows to learn convolutional filters optimised on the task of distinguishing the animals in $G_{\sigma(i-1)}$ based on their local labelling in the global fragment.

    \item Conclude the training according to the conditions listed in~\cref{sec:idcnn_stop}.

    \item Iterate on $\sigma\left(\mathcal{G}\right)$ until $95\%$ of the images stored in the global fragments have been used to train the network.
\end{enumerate}

\paragraph{Accumulation parachute.\label{sec:accumulation_parachute}} After pretraining, we start the accumulation of reference images as in~\cref{sec:protocol_two}, but freezing the parameters of $\mbox{idCNN}_c$ learnt during pretraining along the entire accumulation. Thus, in the first step of this second accumulation we reinitialise only the fully-connected part of the idCNN. With these settings, we apply the accumulation protocol, by updating only the parameters $\Phi$, and starting by considering the global fragment $G_{\sigma(1)}$.

If more than $90\%$ of the images in the global fragments are accumulated during the accumulation, we proceed to the identification of the individual fragments not used for training~\cref{sec:final_identification}. Otherwise, we will repeat the accumulation starting from $G_{\sigma\left(2\right)}$. If the accumulation fails even in this case, we repeat it with $G_{\sigma\left(3\right)}$ as a basis. Finally, we end the deep protocol cascade by selecting the accumulation in which the largest amount of images has been used for traninig and hence, already identified. By using the parameters of the idCNN learnt in the chosen accumulation, we proceed to the identification of the remaining individual fragments.

\begin{rem}
	As pointed out in~\cref{sec:choose_first_global_fragment}, the computation of the distance travelled cannot guarantee the images in $G_{\sigma\left(1\right)}$ to be maximally uncorrelated. Hence it is important, rather than assigning identities with a non-optimal model, to try and learn starting from different global fragments, that could incorporate images whose features are keys to maximise the amount of accumulated global fragments.
\end{rem}

\subsection{Residual identification\label{sec:final_identification}}

After the fingerprint protocol cascade, it is necessary to identify those individual fragments that could not be accumulated, either because they are not included in any global fragment, or they gave a low certainty value during test. We recall that all the individual fragments already accumulated are endowed with both an identity and the $P_1$-vector. See~\cref{sec:individual_fragments_identification_1,eq:P1} for details about the identification of individual fragments during accumulation.

\subsubsection{Non-accumulated images identification}

Let $\mathcal{U} = \Set{F_1, \dots, F_n}$ be the set of individual fragments that are not identified during the protocols described in~\cref{sec:deep_protocol_cascade}. First, we assign an identity to every image $I$ in every $F_i$ in $\mathcal{U}$. To do that, we pass every image through $\mbox{isCNN}\left(\Theta_{\mbox{final}}\right)$, where $\Theta_{\mbox{final}}$ are the parameters learnt during the fingerprint protocol cascade. Then $P_1\left(F_i\right)$ is computed for every $F_i\in\mathcal{U}$ according to~\cref{sec:individual_fragments_identification_1,eq:P1}.

\subsubsection{Identification of non-accumulated individual fragments\label{sec:individual_fragments_identification_P2}}

When assigning the identity to an individual fragment, it is desirable to take advantage of the fact that the same identity cannot be assigned to two fragments that coexist in time. Given an individual fragment $F$, let $\overline{\gamma}\left(F\right) = \Set{\overline{F_1}, \dots, \overline{F_n}}$ be the set of identified individual fragments coexisting with $F$ and not $F$ itself, and such that every $\overline{F_i}\in\gamma\left(F\right)$ is equipped with a $P_1$-vector. We integrate the information coming from the identified fragments coexisting with $F$ by following the approach of \cite[Supporting text, Section 3.1]{Perez-Escudero2014}. We define the probability of the fragment $F$ to be assigned to the identity $i$ as

\begin{equation}\label{eq:P2}
P_2\left(F, i \right) = \frac{P_1\left(F, i \right) \prod_{\overline{\gamma}\left(F\right)}\left( 1 - P_1\left(\overline{F}, i \right)\right)}{\sum_{j=1}^n{P_1\left(F, j\right) \prod_{\overline{F}\in\overline{\gamma}\left(F\right)}\left( 1 - P_1\left(\overline{F}, j \right)\right)} }.
\end{equation}
where $n$ is the total number of animals.

Furthermore, we define the identification certainty as
\begin{equation}\label{eq:certainty_2}
\overline{cert}(F) = \frac{P_2\left(F, a\right)}{P_2\left(F, b\right)},
\end{equation}
where $a$ and $b$ are the indices (identities) that realise the first and second maximum of $P_1\left(F\right)$, respectively.

We compute $P_2\left(F\right) = \left(P_2\left(F, 1\right), \dots, P_2\left(F, n\right)\right)$ and  $\overline{cert}\left(F\right)$ for every individual fragment $F\in U$. We proceed to identify the fragments in $\mathcal{U}$ from higher to lower values of $\overline{cert}$.
For every fragment we assign the identity $\iota = \argmax_i \left( P_2 \left( F, i \right) \right)$. If there are two identities realising the maximum $P_2$, we do not identify the fragment (in the GUI this fragments are indicated with the identity $0$ and black colour). If the fragment $F$ is identified, say with identity $i$, we set $P_1 \left( F, i \right) = 1$ and $P_1 \left( F, j \right) = 0, \forall j \neq i$ . Then, we recompute $P_2 \left( F \right)$ and $\overline{cert}$ for every fragment in $\overline{\gamma}\left(F\right)$. According to \cref{eq:P2}, all the fragments in $\overline{\gamma}\left(F\right)$ will have $P_2 \left( F, i \right) = 0$. This prevents the assignment of the same identity to multiple coexisting individual fragment.

The process is iterated on $\mathcal{U}\setminus\Set{F}$, until all the fragments are either equipped with an identity or unsuitable for identification.

\subsection{Post-processing}
The training/identification protocols and the residual identification assigned an identity to a as large as possible number of individual fragments. The methods involved in the post-processing stage of the algorithm take care of correcting trivial identification mistakes and, thereafter, to identify the individuals involved in crossings.

These processes are described in details in the following sections; here, we provide an intuition concerning the algorithms involved in both of them. It is possible to correct trivial identification errors by considering adjacent individual fragments assigned to the same individual. If the individual has to reach a supernatural speed in order to move from its position at the end of a fragment, to the position corresponding to the beginning of the next one, the identification is assumed to be incorrect. A series of heuristics allows to either assign a new (not necessarily different) identity to the fragments involved in the process, or renounce to their identification.

The idea underlying the identification of crossings is basically an informed interpolation of the individual trajectory. First, we consider a blob associated to a crossing. We workout the identities of each crossing individual by trying to split the blob by successive erosions. If the blob splits in smaller parts (say sub-blobs), we try to link each sub-blob to an already identified individual fragment. To do that, we consider two conditions. On one hand we evaluate the eventual overlapping of the sub-blobs we just obtained with identified, individual blobs segmented either in the next or the previous frame. In case the overlapping strategy fails, we seek individual blobs in adjacent frames with respect to the considered crossing, that can be linked to the sub-blob by using speed-constraints similar to the ones discussed in the previous paragraph.

\subsubsection{Evaluate unrealistic identifications at fragment boundaries}

Individual fragments are defined by considering the overlapping of blobs segmented from consecutive frames (see~\cref{sec:fragmentation}). Let us denote the frame numbers spanned by a fragment $F$ as $[f_s, f_e]$, where $f_s$ is the number of the frame from which the first blob associated to $F$ has been segmented. We say that two individual fragments $F_1$ and $F_2$ are \textit{consecutive} if they share the same identity, say $i$, and $f_{1e} < f_{2s}$. We aim at evaluating the goodness of the identification of such fragments by comparing a model of the stereotypical speed of the individuals in the video, with the speed that the individual $i$ needs to reach to travel from its position in $f_{1e}$ to its new position in $f_{2s}$.

\paragraph{Computation of the stereotypical speed:} The stereotypical speed is computed as follows by considering the speed of the animals in every individual fragment:
\begin{enumerate}
  \item For every individual fragment $F$, let $\left(b_1,\dots, b_n\right)$ be the blobs collected in $F$, and $\left(c_1, \dots, c_n\right)$ their centroids.

  \item We compute the speed of the animal in $F$ by considering the distance in pixels between subsequent centroids. Namely, $v_i = d\left(c_i, c_{i+1}\right)$ for $i\in\Set{1,\dots, n-1}$.

  \item We define $v_{\mbox{max}} = P_{99}\left(V\right)$, where $V$ collects the speeds computed from every individual fragment $F$.
\end{enumerate}

\paragraph{Evaluation of consecutive fragments:}
We set to immutable the identity of all the fragments that have either been identified during the deep protocol cascade, or whose identity has been assigned during the residual identification with $\max_i(P_2(F)) \geq 0.9$.

Let $F_1$ and $F_2$ be the consecutive fragments described above. The speed at the boundary $s_{F_1\rightarrow F_2} = \frac{d(c_{1e}, c_{2s})}{f_{2s} - f_{1e}}$  needed to connect the two fragment is \textit{realistic} if $s_{F_1\rightarrow F_2}\leq 2v_{\mbox{max}}$. In order to test and correct for unrealistic connecting speed, we proceed as follows:
We iterate on the collection of individual fragments by separating them into two subsets. First we consider the individual fragments whose last frame is less than the core of the first global fragment used for training (see~\cref{sec:choose_first_global_fragment}), then the others. Let us consider a general individual fragment $F$ spanning frame numbers $[f_s, f_e]$. We check if there exist fragments $F_p$ and $F_n$ sharing the identity with $F$ and defined either in the past or in the future.
If $F_p$ and $F_n$ do not exists, we proceed with the iteration. Otherwise, we evaluate the boundary speeds $s_{F_p\rightarrow F}$ and $s_{F\rightarrow F_n}$. We distinguish the following cases:
\begin{enumerate}
  \item $s_{F_p\rightarrow F} > 2v_{\mbox{max}}$ and $s_{F\rightarrow F_n} > 2v_{\mbox{max}}$: If the identity of $F_p$ or $F_n$ is fixed or neither $F_p$'s consecutive previous fragment, nor $F_n$'s consecutive next fragments are unrealistic, we set $F$ to be reidentified.

  \item $s_{F_p\rightarrow F} > 2v_{\mbox{max}}$ and $s_{F\rightarrow F_n} \leq 2v_{\mbox{max}}$:
  If the identity of $F_p$ is fixed $F$ is set to be reidentified. Otherwise $F_p$.

  \item Symmetrically in the case $s_{F_p\rightarrow F} \leq 2v_{\mbox{max}}$ and $s_{F\rightarrow F_n} > 2v_{\mbox{max}}$.
\end{enumerate}

\paragraph{Reidentification of unrealistic consecutive fragments:} Let $F$ be an individual fragment to be reidentified. We compute the set $A$ of available identities by considering the identities of the individual fragments coexisting with $F$, and including the identity of $F$ itself. If $|A| = 1$, we assign the only available identity to $F$. Otherwise we proceed by calculating:
\begin{enumerate}
  \item The subset $S\subseteq A$ of available identities that would not imply unrealistic boundary speeds given $F$.
  \item $Q = \Set{i\in A \mbox{ s.t. }P_2\left(F, i \right) > \rho(F)}$, where
  \[
  \rho(F) =\begin{cases}
    \frac{1}{|F|}\mbox{ if }|F|>1\\
    \frac{1}{n}\mbox{ otherwise}
  \end{cases},
  \]
  where $n$ is the number of tracked animals.
  \item The set $C = Q\cap A$ of candidate identities, \ie the set of identities that do not create duplications if assigned to $F$ and are at the same time acceptable with respect to both $P_2(F)$ and the speed model.
\end{enumerate}

\noindent By considering the set $C$ just defined we have:
\begin{enumerate}
  \item $|C| = 0$: No identities are available, thus $F$ is not identified.
  \item $C = \Set{i}$: We identify $F$ with $i$.
  \item $|C| > 1$: $F$ is identified by considering the identity $i\in C$ realising the minimum boundary speed.
\end{enumerate}

\subsubsection{Crossing identification\label{sec:crossing_identification}}
Single individuals in a crossing are identified according to a python reimplementation of the algorithm described in~\cite[Supporting Text, Section 2.12]{Perez-Escudero2014}. See~\href{http://idtracker.ai/postprocessing/}{idtracker.ai/postprocessing/} for the documentation and the source code of the algorithm.

\subsection{Output}

In this section we discuss the final outputs of the algorithm: An estimation of the tracking accuracy will warn the user in case the algorithm could not proceed smoothly in the tracking process. The files containing the trajectories of each individual are saved and made available to the user.

\subsubsection{Estimated accuracy\label{sec:estimated_accuracy}}
Let $\mathcal{I}$ be the set of all identified individual fragments, and $N = \sum_{F\in\mathcal{I}} |F|$ the total number of images in such fragments. We estimate the overall accuracy of the algorithm as
\[
\mathcal{A} = \frac{\sum_{F\in\mathcal{I}} P_2(F, i) \cdot |F|}{N},
\]
where $i$ is the identity assigned by the algorithm to the fragment $F$.

\subsubsection{Individual trajectories}
The algorithms outputs two individual trajectories files. One generated by considering the identification of individual images; the second by including the identification of the individual during crossings (see~\cref{sec:crossing_identification}).

Both are organised as matrices with shape $\mbox{number of frames}\times\mbox{number of individuals}\times 2$, where the last two components are the position of the centroid of each individual in pixel coordinate, with respect to the entire frame.

\section{Human validation\label{sec:human_validation}}

After a video has been tracked, idtracker.ai provides an estimate of its own accuracy (see~\cref{sec:estimated_accuracy}). Human validation is necessary to evaluate the goodness of the automatic accuracy assessment, notice recurrent inaccurate identifications, and evaluate the limit conditions in which the tracking system can work (\eg fitness of the setup, suitability of the recording conditions and quality of the images).

We recall that the identity of an individual is maintained throughout individual fragments, thus a misidentification can happen only after a crossing or an occlusion. Notice that here a bad segmentation of the images (see~\cref{sec:segmentation}) counts as an occlusion. Hence, the optimal validation would consist in evaluating that the identities of the animals before and after every crossing is conserved. Identities are assigned for the first time when labelling the individual fragments of the first global fragment used to train the idCNN. Hence, only by starting the validation from that global fragment, one can be sure that no switch of identities between two or more individuals occurred.

When dealing with large groups or particularly long videos, the validation of all the crossings is extremely costly. For this reason, we provide two procedures to facilitate this process. On the one hand, a global validation graphic interface allows to easily check the goodness of the identification of all the individuals in a segment of the video, correct their identities and compute the accuracy of the identification by considering the user-generated ground truth. On the other hand, an individual validation procedure allows to select a specific animal and follow it throughout the video. All the crossings, or occlusions that do not involve that individual are ignored, allowing a fast validation in long segments of the video.

\subsection{Global validation\label{sec:global_validation}}
Starting from the core of the first global fragment (see~\cref{sec:global_fragments}), we manually check that in every crossing all the identities of the animal involved in are maintained. Corrected identities are stored. After providing the segment $S = \left(\mbox{start - end}\right)$ on which validation has been performed, we compute the following accuracy indices. Let $\mathcal{I}_S$ be the total number of individual images validated.
\begin{enumerate}[1.]
	\item \textbf{Accuracy during protocol cascade}: Number of images correctly identified during the fingerpint protocol cascade, over the total number of individual images used to train the idCNN in $S$.

  \item \textbf{Accuracy}: Number of images correctly identified over $\mathcal{I}_S$.

  \item \textbf{Percentage of non-identified images}: Number of images non-identified, over $\mathcal{I}_S$.

  \item \textbf{Percentage of misidentified images}: Number of images misidentified, over $\mathcal{I}_S$.

\end{enumerate}

\subsection{Individual validation\label{sec:individual_validation}}
Individual validation is performed by considering a single individual at a time, and always proceeding from the core of the first global fragment used in the protocol cascade to the previous or the future frames. When validating the individual assigned to the identity $\iota$, we are interested only in crossings and occlusions in which it is involved. In this way, the validation is much faster and it is possible to control the quality of the identification in a wider timespan. After the correction of the misidentified images, we compute the accuracy in the assignment of $\iota$ by considering the number of correctly identified images, over the number of total images representing the individual.

\clearpage
\mbox{}

\nomenclature[1a]{$A = \Set{a_1,\dots, a_n}$}{Set collecting the elements $a_1, \dots, a_n$.}
\nomenclature[2a]{$\vert A\vert$}{Number of elements in $A$.}
\nomenclature[3a]{$a\in A$}{The element $a$ belongs to the set $A$.}
\nomenclature{$A\cap B$}{Sets intersection: Set of the elements shared by $A$ and $B$.}
\nomenclature{$A\cup B$}{Sets union: Union of the elements of $A$ and $B$.}
\nomenclature{$A\setminus B$}{Difference between sets: Set of the elements of $A$ that are not elements of $B$.}
\nomenclature{s.t.}{Such that.}

\printnomenclature[8em]
%%%%%%%%%%%%%%%%%%%%%%%%%%%%%%%%%%%%%%%%%%%%%%%%%%%%%%%%%%%%%%%%%%%%%%%%%%%%%%%%%%%%%%%%%%%%%%%%%

\end{appendices}

\addcontentsline{toc}{section}{References}
\bibliographystyle{unsrt}
\bibliography{idtracker_bib}

\end{document}